\RequirePackage[hyphens]{url}
\RequirePackage[prologue,table]{xcolor}

\documentclass[acmsmall,screen,authorversion]{acmart}
\makeatletter
\def\mdseries@tt{m}
\makeatother
\usepackage{fvextra}

\setcopyright{rightsretained}
\acmJournal{PACMHCI}
\acmYear{2021}
\acmVolume{5}
\acmNumber{CSCW2}
\acmArticle{431}
\acmMonth{10}
\acmPrice{}
\acmDOI{10.1145/3479575}

\usepackage{graphicx}
\graphicspath{{figures/}}
\DeclareGraphicsExtensions{.pdf,.jpeg,.png}

\usepackage{amsmath}
\usepackage{amssymb}           \usepackage{amsthm}
\usepackage[hyphens]{url}

\usepackage{array}
\usepackage{balance}           \usepackage{bbding}            \usepackage{bm}                \usepackage{booktabs}          \usepackage{caption}           \usepackage{enumitem}
\usepackage{fancyvrb}          \usepackage{lipsum}            \usepackage{makecell}          \usepackage{subcaption}        \usepackage{tabularx}          \usepackage{tabulary}          \usepackage{textcomp}          \usepackage{tikz}
\usepackage{xspace}

\usetikzlibrary{shapes,arrows.meta,positioning,decorations.text}

\usepackage[frozencache,cachedir=_minted-main,newfloat=true]{minted}

\usepackage{color}
\usepackage{colortbl}
\usepackage[prologue,table]{xcolor}

\usepackage{mscommon}
\newtheorem{definition}{Definition}

\usepackage{algorithm2e}
\RestyleAlgo{ruled}
\LinesNumbered

\SetCommentSty{mycommentstyle}
\SetAlCapSkip{1ex}
\DontPrintSemicolon
\SetKwInOut{Input}{input}
\SetKwInOut{Output}{output}
\SetKwInOut{Params}{params}

\setlist[itemize]{leftmargin=*, itemindent=0pt, topsep=2pt, itemsep=2pt, parsep=2pt, partopsep=2pt}
\setlist[enumerate]{leftmargin=*, labelindent=0pt, topsep=2pt, itemsep=2pt, parsep=2pt, partopsep=2pt}

\usepackage{varioref}
\usepackage{hyperref}
\usepackage[capitalise]{cleveref}
\newcommand{\mytitle}{Enabling Collaborative Data Science Development with the Ballet Framework}

\newcommand{\featurefunction}{feature function\xspace}
\newcommand{\featurefunctions}{feature functions\xspace}

\newcommand{\Featurefunctions}{Feature functions\xspace}

\newcommand{\D}{\mathcal{D}}
\newcommand{\F}{\mathcal{F}}
\newcommand{\X}{\mathcal{X}}
\newcommand{\Y}{\mathcal{Y}}
\newcommand{\V}{\mathcal{V}}
\newcommand{\inline}[1]{{\small\fontfamily{cmtt}\selectfont{#1}}}

\newcommand{\ie}{i.e.,\xspace}
\newcommand{\Ballet}{Ballet\xspace}
\newcommand{\Assemble}{Assembl\'{e}\xspace}
\newcommand{\pci}{\emph{predict-census-income}\xspace}

\newcommand{\myrevision}[1]{#1}

\newcommand{\balletsiteurl}{https://ballet.github.io}
\newcommand{\balleturl}{https://github.com/ballet/ballet}

\newcommand{\balletcensusurl}{https://github.com/ballet/predict-census-income}
\newcommand{\assembleurl}{https://github.com/ballet/ballet-assemble}
\newcommand{\balletboturl}{https://github.com/ballet/ballet-bot}
\newcommand{\replicationurl}{https://github.com/micahjsmith/ballet-cscw-2021}

\makeatletter
	\newcommand{\algorithmicreturn}{\textbf{return}}
	\newcommand{\RETURN}{\ALC@it\algorithmicreturn{} \ }

\makeatother

\begin{document}

\title{\mytitle}

\author{Micah J. Smith}
\email{micahs@mit.edu}
\affiliation{  \institution{Massachusetts Institute of Technology}
  \city{Cambridge}
  \state{Massachusetts}
  \country{USA}
}

\author{J\"{u}rgen Cito}
\email{juergen.cito@tuwien.ac.at}
\affiliation{  \institution{TU Wien}
  \city{Vienna}
  \country{Austria}
}
\affiliation{  \institution{Massachusetts Institute of Technology}
  \city{Cambridge}
  \state{Massachusetts}
  \country{USA}
}

\author{Kelvin Lu}
\email{kellu1997@gmail.com}
\affiliation{  \institution{Massachusetts Institute of Technology}
  \city{Cambridge}
  \state{Massachusetts}
  \country{USA}
}

\author{Kalyan Veeramachaneni}
\email{kalyanv@mit.edu}
\affiliation{  \institution{Massachusetts Institute of Technology}
  \city{Cambridge}
  \state{Massachusetts}
  \country{USA}
}

\renewcommand{\shortauthors}{Micah J. Smith et al.}

\begin{abstract}
    While the open-source software development model has led to successful large-scale collaborations in building software systems, data science projects are frequently developed by individuals or small teams. We describe challenges to scaling data science collaborations and present a conceptual framework and ML programming model to address them. We instantiate these ideas in \Ballet, the first lightweight framework for collaborative, open-source data science through a focus on feature engineering, and an accompanying cloud-based development environment. Using our framework, collaborators incrementally propose feature definitions to a repository which are each subjected to software and ML performance validation and can be automatically merged into an executable feature engineering pipeline. We leverage \Ballet to conduct a case study analysis of an income prediction problem with 27 collaborators, and discuss implications for future designers of collaborative projects.
\end{abstract}



\maketitle

\section{Introduction}
\label{sec:intro}

The open-source software development model has led to successful, large-scale collaborations in building software libraries, software systems, chess engines, scientific analyses, and more \cite{raymond1999cathedral, linux, gnu, stockfish, bos2007shared}. However, data science, and in particular, predictive machine learning (ML) modeling, has not similarly benefited from this development paradigm. Predictive modeling projects --- where the output of the project is not general-purpose software but rather a specific trained model and library capable of serving predictions for new data instances --- are rarely developed in open-source collaborations, and when they are, they generally have orders of magnitude fewer contributors (\Cref{tab:project-sizes} and \cite{choi2017characteristics}).

There is great potential for large-scale, collaborative data science to address societal problems through community-driven analyses of public datasets \cite{choi2017characteristics, hou2017hacking}. For example, the Fragile Families Challenge tasked researchers and data scientists with predicting outcomes, including GPA and eviction, for a set of disadvantaged children and their families \cite{salganik2020measuring}, and \textit{crash-model} is an application for predicting car crashes and thereby directing safety interventions \cite{crashmodel}. Such projects, which involve complex and unwieldy datasets, attract scores of interested citizen scientists and developers whose knowledge, insight, and intuition can be significant if they are able to contribute and collaborate.

To make progress toward this outcome, we must first better understand the capabilities and challenges of collaborative data science as projects scale beyond small teams. We make two main contributions toward this goal.

First, we ask, \emph{can we support larger collaborations in predictive modeling projects by applying successful open-source development models?} In this work, we show that we can successfully adapt and extend the pull request development model to support collaboration during important steps within the data science process \myrevision{by introducing} a new development workflow and ML programming model. Our approach is based on decomposing steps in the data science process into modular data science ``patches'' that can then be intelligently combined, representing objects like ``feature definition,'' ``labeling function,'' or ``slice function.'' Prospective collaborators work in parallel to write patches and submit them to a shared repository. Our framework provides the underlying functionality to support interactive development, test and merge high-quality contributions, and compose the accepted contributions into a single product. Projects built with \Ballet are structured by these modular patches, yielding additional benefits including reusability, maintainability, reproducibility, and automated analysis. Our lightweight framework does not require any computing infrastructure beyond that which is freely available in open-source software development.

While data science and predictive modeling have many steps, we focus on feature engineering as an important one that could benefit from a more collaborative approach. We thus instantiate these ideas in \emph{\Ballet},\footnote{\url{\balletsiteurl}, \url{\balleturl}} the first lightweight software framework for collaborative data science that supports collaborative feature engineering on tabular data. Developers can use Ballet to grow a shared feature engineering pipeline by contributing feature definitions, which are each subjected to software and ML performance validation. Together with \emph{\Assemble} --- a data science development environment customized for \Ballet~--- even novice developers can contribute to large collaborations.

Second, we seek to better understand the opportunities and challenges in large open-source data science collaborations. Although much research has focused on elucidating diverse challenges involved in data science development \cite{chattopadhyay2020wrong, yang2018grounding, subramanian2020casual, sculley2015hidden, choi2017characteristics, zhang2020how, muller2019how}, little attention has been given to large collaborations, partly due to the lack of real-world examples available for study. Leveraging \Ballet as a probe, we create and conduct an analysis of \pci, a collaboration to predict personal income through engineering features from raw individual survey responses to the U.S. Census American Community Survey (ACS). We use a mixed-method software engineering case study approach to study the experience of 27 developers collaborating on this task, focusing on understanding the experience and performance of participants from varying backgrounds, the characteristics of collaboratively developed feature engineering code, and the performance of the resulting model compared to alternative approaches. The resulting project is one of the largest ML modeling collaborations on GitHub, and outperforms both state-of-the-art tabular AutoML systems and independent data science experts. We find that both beginners and experts (in terms of their background in software development, data science, and the problem domain) can successfully contribute to such projects and that domain expertise in collaborators is critical. We also identify themes of goal clarity, learning by example, distribution of work, and developer-friendly workflows as important touchpoints for future design and research in this area.

\begin{table}[t]
  \centering
  \caption[]{The number of unique contributors to large open-source collaborations in either software engineering or predictive machine learning modeling. ML modeling projects that are developed in open-source have orders of magnitude fewer contributors. (The methodology is described in \Cref{app:largest-collaborations}.)}
  \label{tab:project-sizes}
      \begin{tabular}{lclc}
      \toprule
      \multicolumn{2}{l}{Software engineering} & \multicolumn{2}{l}{ML modeling} \\
      \midrule
                  torvalds/linux & 20,000+ & tesseract-ocr/tesseract & 130 \\
      DefinitelyTyped/DefinitelyTyped & 12,600+ & CMU-PCL/openpose & 79 \\
      Homebrew/homebrew-cask & 6,500+ & deepfakes/faceswap & 71 \\
      ansible/ansible & 5,100+ & JaidedAI/EasyOCR & 62 \\
      rails/rails & 4,300+ & ageitgey/face\_recognition & 43 \\
      gatsbyjs/gatsby & 3,600+ & \pci (this work) & 27 \\
      helm/charts & 3,400+ & microsoft/CameraTraps & 21 \\
      rust-lang/rust & 3,000+ & Data4Democracy/drug-spending & 21 \\
                  \bottomrule
    \end{tabular}  \end{table}

The remainder of this paper proceeds as follows. We first provide background on data science, collaborative data work, and open-source software development. We then describe a conceptual framework for collaboration in data science projects in \Cref{sec:conceptual}, which we then apply to create the \Ballet framework in \Cref{sec:overview}. We next describe key components of \Ballet{'s} support for collaborative feature engineering in \Cref{sec:fteng}, such as the feature definition abstraction, feature engineering pipeline abstraction, feature validation algorithms, and \Assemble development environment. Next, we introduce the method and procedures for the \pci case study in \Cref{sec:user-studies} and analyze the results in \Cref{sec:results}. We discuss our work and results in \Cref{sec:discussion}, including future directions for designers and researchers in collaborative data science frameworks.

\section{Background and related work}
\label{sec:background}

\subsection{Data science and feature engineering}
\label{sec:background:data-science-and-feature-engineering}

The increasing availability of data and computational resources has led many organizations to turn to data science, or a data-driven approach to decision-making under uncertainty. Consequently, researchers have studied data science work practices on several levels, and the data science workflow is now understood as a complex, iterative process that includes many stages and steps. The stages can be summarized as Preparation, Modeling, and Deployment \cite{muller2019how, wang2019humanai} and encompass other smaller steps such as data cleaning and labeling, feature engineering, model development, monitoring, and analyzing bias. Within the larger set of data science workers involved in this process, we use \emph{data science developers} to refer to those who write code in data science projects.

Within this broad setting, the step of feature engineering holds special importance in some applications. Feature engineering --- also called feature creation, development, or extraction --- is the process through which data science developers write code to transform raw variables into feature values that can be used as input to a machine learning model. (This process is sometimes grouped with data cleaning and preparation steps, as in \citealt{muller2019how}.) Features form the cornerstone of many data science tasks, including not only predictive ML modeling, in which a learning algorithm finds predictive relationships between feature values and an outcome of interest, but also causal modeling through propensity score analysis, clustering, business intelligence, and exploratory data analysis. Practitioners and researchers have widely acknowledged the importance of engineering good features for modeling success, particularly in predictive modeling \cite{domingos2012few, anderson2013brainwash, veeramachaneni2014feature}.

Before we continue discussing feature engineering, we introduce some terminology that we will use throughout this paper. A \emph{\featurefunction} is a transformation applied to raw variables that extracts \emph{feature values}, or measurable characteristics and properties of each observation. A \emph{feature definition} is source code written by a developer to create a \featurefunction.\footnote{Any of these terms may be referred to as ``features'' in other settings, but we make a distinction between the source code, the transformation applied, and the resulting values. In cases where this distinction is not important, we may also use ``feature.''} If many \featurefunctions are created, they can be collected into a single \emph{feature engineering pipeline} that executes the computational graph made up of all of the \featurefunctions and concatenates the result into a \emph{feature matrix}.

In an additional step in ML systems, feature engineering is increasingly augmented by applications like feature stores and feature management platforms to help with critical functionality like feature serving, curation, and discovery \cite{michelangelo, wooders2021feature}.

Though there have been attempts to automate the feature engineering process in certain domains, including relational databases and time series analysis \cite{kanter2015deep, khurana2016cognito, christ2018time, katz2016explorekit}, it is widely accepted that in many areas that involve large and complex datasets, like health and business analytics, human insight and intuition are necessary for success in feature engineering \cite{domingos2012few, smith2017featurehub, wagstaff2012machine, veeramachaneni2014feature, bailis2020humans}. Human expertise is invaluable for understanding the complexity of a dataset, theorizing about different relationships, patterns, and representations in the data, and implementing these ideas in code in the context of the machine learning problem. \citet{muller2019how} observe that ``feature extraction requires an interaction of domain knowledge with practices of design-of-data.'' As more people become involved in this process, there is a greater chance that impactful ``handcrafted'' feature ideas will be expressed; automation can be a valuable supplement to manual development.
Indeed, \citet{smith2017featurehub} introduce a feature engineering platform in which contributors log in to a cloud platform and submit source code directly to a machine learning backend server.

In this paper, we build on understanding of the importance of human interaction within the feature engineering process by creating a workflow that supports collaboration in feature engineering as a component of a larger data science project. \Ballet takes a lightweight and decentralized approach suitable for the open-source setting, an integrated development environment, and a focus on modularity and supporting collaborative workflows.

\subsection{Collaborative and open data work}
\label{sec:background:collaborative-and-open-data-work}

Just as we explore how multiple human perspectives enhance feature engineering, there has been much interest within the HCI and CSCW communities in achieving a broader understanding of collaboration in data work. For example, within a wider typology of \emph{collaboratories} (collaborative organizational entities), \citet{bos2007shared} study both community data systems and open community contribution systems, such as the Protein Databank and Open Mind Initiative. \citet{zhang2020how} show that data science workers in a large company are highly collaborative in small teams, using a plethora of tools for communication, code management, and more. Teams include workers in many roles such as researchers, engineers, domain experts, managers, and communicators \cite{muller2019how}, and include both experts and non-experts in technical practices \cite{middleton2020data}. In an experiment with a prototype machine learning platform, \citet{smith2017featurehub} show that 32 data scientists made contributions to a shared feature engineering project and that a model using all of their contributions outperformed a model from the best individual performer. Functionalities including a feature discovery method and a discussion forum helped data scientists learn how to use the platform and avoid duplicating work. Often, data science workers in teams coordinate around a shared work product, such as a data science pipeline, like in the MIDST project \cite{crowston2019sociotechnical}, which introduces a conceptual framework for collaboration in a shared data science project around the concepts of visibility, combinability, and genre. We expand on this body of work by extending the study of collaborative data work to predictive modeling and feature engineering and by using the feature engineering pipeline, as a shared work product, to coordinate collaborators at a larger scale than previously observed.

One finding in common in previous studies is that data science teams are usually small, with six or fewer members \cite{zhang2020how}. There are a variety of explanations for this phenomenon in the literature.  Technical and non-technical team members may speak ``different languages'' \cite{hou2017hacking}. Different team members may lack common ground while observing project progress and may use different performance metrics \cite{mao2019how}. Individuals may be highly specialized, and the lack of a true ``hub'' role on teams \cite{zhang2020how} along with the use of synchronous communication forms like telephone calls and in-person discussion \cite{choi2017characteristics} make communication challenges likely as teams grow larger. One possible implication of this finding is that, in the absence of other tools and processes, human factors of communication, coordination, and observability make it challenging for teams to work well at scale. Difficulties with validation and curation of feature contributions presented challenges for \citet{smith2017featurehub}, which points to the limitations of existing feature evaluation algorithms. Thus, algorithmic challenges may complement human factors as obstacles to scaling data science teams. However, additional research is needed into the question of why data science collaborations are not larger. We provide a starting point through a case study analysis in this work.

Moving from understanding to implementation, other approaches to collaboration in data science work include crowdsourcing, synchronous editing, and competition. Unskilled crowd workers can be harnessed for feature engineering tasks within the Flock platform, such as by labeling data to provide the basis for further manual feature engineering \cite{cheng2015flock}. Synchronous editing interfaces, like those of Google Colab and others for computational notebooks \cite{garg2018fabrik,jupyter,wang2019how}, facilitate multiple users to edit a machine learning model specification, typically targeting pair programming or other very small groups. In our work, we explore \emph{different-time, different-place} collaboration \cite{shneiderman2016designing} in an attempt to move beyond the limitations of small group work. A form of collaboration is also achieved in data science competitions like the KDD Cup, Kaggle, and the Netflix Challenge \cite{bennett2007netflix} and using networked science hubs like OpenML \cite{vanshoren2013openml}. While these have led to state-of-the-art modeling performance, there is no natural way for competitors to systematically integrate source code components into a single shared product. In addition, individual teams formed in competitions hosted on Kaggle are small, with the mean team having 2.6 members and 90\% of teams having four or fewer members, similar to other types of data science teams as discussed above.\footnote{Author's calculation from \citet{metakaggle} of all teams with more than one member.}

Closely related is \emph{open data analysis} or \emph{open data science}, in which publicly available datasets are used by ``civic hackers'' and other technologists to address civic problems, such as visualizations of lobbyist activity and estimates of child labor usage in product manufacturing \cite{choi2017characteristics}. Existing open data analysis projects involve a small number of collaborators (median of three) and make use of synchronous communication \cite{choi2017characteristics}. A common setting for open data work is hackathons, during which volunteers collaborate with non-profit organizations to analyze their internal and open data. \citet{hou2017hacking} find that civic data hackathons create actionable outputs and improve organizations' data literacy, relying on ``client teams'' to prepare data for analysis during the events and to broker relationships between participants. Looking more broadly at collaborative data work in open science, interdisciplinary collaborations in data science and biomedical science are studied in \citet{mao2019how}, who find that readiness of a team to collaborate is influenced by its organizational structures, such as dependence on different forms of expertise and the introduction of an intermediate broker role. In our work, we are motivated by the potential of open data analysis, but focus more narrowly on data science and feature engineering.

\subsection{Open-source development}
\label{sec:background:open-source-development}

The \emph{open-source model} for developing software has been adopted and advanced by many individuals and through many projects \cite{raymond1999cathedral}. In the open-source model, projects are developed publicly and source code and other materials are freely available on the internet; the more widely available the source code, the more likely it is that a contributor will find a defect or implement new functionality (``with enough eyes, all bugs are shallow''). With freely available source code, open-source projects may attract thousands of contributors: developers who fix bugs, contribute new functionality, write documentation and test cases, and more. With more contributors comes the prospect of conflicting patches, leading to the problem of \emph{integration}.
In order to support open-source developers, companies and organizations have made a variety of lightweight infrastructure and developer tooling freely available for this community, such as build server minutes and code analysis tools.

Closely associated with the open-source model is the \emph{open-source software development process}, exemplified by the \emph{pull-based development model} (or \emph{pull request model}), a form of distributed development in which changes are pulled from other repositories and merged locally. As implemented on the social code platform GitHub, developers fork a repository to obtain their own copy and make changes independently; proposed changes (pull requests) are subject to discussions and code reviews in context and are analyzed by a variety of automated tools. The pull request model has been successful in easing the challenges of integration at scale and facilitating massive software collaborations.

As of 2013, 14\% of active repositories on GitHub used pull requests. An equal proportion used shared repositories without pull requests, while the remainder were single-developer projects \cite{gousios2014exploratory}. Pull request authors use contextual discussions to cover low-level issues but supplement this with other channels for higher-level discussions \cite{gousios2016work}. Pull request integrators play a critical role in this process but can have difficulty prioritizing contributions at high volume \cite{gousios2015work}.
Additional tooling has continued to grow in popularity partly based on these observations. Recent research has visited the use of modern development tools like continuous integration \cite{vasilescu2015quality, zhao2017impact, vasilescu2014continuous}, continuous delivery \cite{schermann2016quality}, and crowdsourcing \cite{latoza2016crowdsourcing}.
In this work, we specifically situate data science development within the open-source development process and explore what changes and enhancements are required for this development model to meet the needs of developers during a collaboration.

\subsection{Testing and end-to-end ML}
\label{sec:background:testing-and-end-to-end-ml}

As part of our framework, we discuss the use of testing in continuous integration to validate contributions to data science pipelines. Other research has also explored the use of continuous integration in machine learning. \citet{renggli2019continuous} investigate practical and statistical considerations arising from testing conditions on overall model accuracy in a continuous integration setting. Specific models and algorithms can be tested \cite{grosse2014testing} and input data can be validated directly \cite{hynes2017data, breck2019data}. Testing can also be tied to reproducibility in ML research \cite{ross2018refactoring}. We build on this work by designing and implementing the first system and algorithms that conduct ML testing at the level of individual feature definitions.

Feature engineering is just one of many steps involved in data science. Other research has looked at the entire endeavor from a distance, considering the end-to-end process of delivering a predictive model from some initial specification. Automated machine learning (AutoML) systems like AutoBazaar, AutoGluon, and commercial offerings from cloud vendors \cite{smith2020machine, erickson2020autogluon} can automatically create predictive models for a variety of ML tasks. A survey of techniques used in AutoML, such as hyperparameter tuning, model selection, and neural architecture search, can be found in \citet{yao2019taking}. On the other hand, researchers and practitioners are increasingly realizing that AutoML does not solve all problems and that human factors such as design, monitoring, and configuration are still required \cite{cambronero2020ams, xin2021whither, wang2019atmseer, wang2021how}. In our experiments, we use an AutoML system to evaluate the performance of different feature sets without otherwise incorporating these powerful techniques into our framework.

\section{Conceptual framework}
\label{sec:conceptual}

Having noticed the success of open-source software development along with the challenges in collaborative data science, we set out to understand whether these two paradigms could complement each other, as well as to better grasp the current state of large-scale collaboration in data science. In this section, we describe the formation of key design concepts that underlie the creation of \Ballet. Our methods reflect an iterative and informal design process that played out over the time we have worked on this problem as well as through two preliminary user studies (\Cref{sec:user-studies:preliminary-studies}).

The pull request model (\Cref{sec:background:open-source-development}) has been particularly successful in enabling integration of proposed changes in shared repositories, and is already used for well over one million shared repositories on GitHub \cite{gousios2015work, gousios2016work}. We informally summarize this development model using the concepts of \emph{product}, \emph{patch}, and \emph{acceptance procedure}. A software artifact is created in a shared repository (product). An improvement to the product is provided in a standalone source code contribution proposed as a pull request (patch). Not every contribution is worthy of inclusion, so high-quality and low-quality contributions must be distinguished (acceptance procedure). If accepted, the pull request can be merged.

Given the success of open-source development processes like the pull request model, we ask: \emph{Can we apply the pull request model to data science projects in order to collaborate at a larger scale?}

\subsection{Challenges}
\label{sec:conceptual:challenges}

When we set out to apply the pull request model to data science projects, we found the model was not a natural fit, and discovered key challenges to address, which we describe here. As people embedded in data science work, we built on our own experience developing and researching feature engineering pipelines and other data science steps from a machine learning perspective.  We also uncovered and investigated these challenges in preliminary user studies with prototypes of our framework (\Cref{sec:user-studies}).

We synthesize these challenges in the context of the literature on collaborative data work and machine learning workflows. Previous work outside the context of open-source development has identified challenges in communication, coordination, observability, and algorithmic aspects (\Cref{sec:background:collaborative-and-open-data-work}). In addition, \citet{brooks1995mythical} observed that the number of possible direct communication channels in a collaborative software project scales quadratically with the number of developers. As a result, at small scales, data science teams may use phone calls or video chats, with 74\% communicating synchronously and in person \cite{choi2017characteristics}. At larger scales, like those made possible by the open-source development process, communication can take place more effectively through coordination around a shared work product, or through discussion threads and chat rooms.

Ultimately, we list four challenges for a collaborative framework to address. While not exhaustive, this list comprises the challenges we mainly focus on in this work, though we review and discuss additional ones
in \Cref{sec:background,sec:discussion}.

\begin{enumerate}[label={C\arabic*}]

  \item \hypertarget{C1}{} \textit{Task management.} Working alone, data science developers often write end-to-end scripts that prepare the data, extract features, build and train model models, and tune hyperparameters \cite{subramanian2020casual, rule2018exploration, muller2019how}. How can this large task be broken down so that all collaborators can coordinate with each other and contribute without duplicating work?

  \item \hypertarget{C2}{} \textit{\myrevision{Tool mismatch.}} Data science developers are accustomed to working in computational notebooks and have varying expertise with version control tools like git \cite{subramanian2020casual, chattopadhyay2020wrong, kery2018story, rule2018exploration}. How can these workflows be adapted to use a shared codebase and build a single product?

  \item \hypertarget{C3}{} \textit{Evaluating contributions.} Prospective collaborators may submit code to a shared codebase. Some code may introduce errors or decrease the performance of the ML model \cite{smith2017featurehub, renggli2019continuous, karlas2020building, kang2020model}. How can code contributions be evaluated?
  
  \item \hypertarget{C4}{} \textit{\myrevision{Maintaining infrastructure.}} Data science requires careful management of data and computation \cite{sculley2015hidden, smith2017featurehub}. Will it be necessary to establish shared data stores and computing infrastructure? Would this be expensive and require significant technical and DevOps expertise? Is this appropriate for the open-source setting?

\end{enumerate}

\begin{table}[!t]
    \renewcommand{\arraystretch}{1.5}
    \centering
    \caption{Addressing challenges in collaborative data science development by applying our design concepts in the \Ballet framework.}
    \label{tab:concept-mapping}
    \resizebox{\linewidth}{!}{    \begin{tabulary}{\linewidth}{p{0.30\linewidth} L L}
        \toprule
        \emph{challenge}
            & \emph{design concept}
            & \emph{components of \Ballet} \\
        \midrule
        task management \newline (C\myrevision{1})
            & data science patches (D\myrevision{1})
            & feature definition abstraction (\ref{sec:fteng:feature-definitions}),\newline
            feature engineering language (\ref{app:fteng}),\newline
              patch development in \Assemble (\ref{sec:overview:collaborators})\\
        \myrevision{tool mismatch} \newline (C\myrevision{2})
            & data science products in open-source workflows (D\myrevision{2})
            & feature engineering pipeline abstraction (\ref{sec:fteng:feature-engineering-pipelines}),\newline
                patch contribution in \Assemble (\ref{sec:overview:collaborators}),\newline
                CLI for project administration (\ref{sec:overview:maintainer}) \\
        evaluating contributions \newline (C3)
            & software and statistical acceptance procedures (D3)
            & feature API validation (\ref{sec:fteng:testing:feature-testing}),\newline
              streaming feature definition selection (\ref{sec:fteng:testing:ml-performance-validation}),\newline
              continuous integration (\ref{sec:overview:maintainer}),\newline
              Ballet Bot (\ref{sec:overview:maintainer}) \\
        \myrevision{maintaining infrastructure} \newline (C4)
            & \myrevision{decentralized development} (D4)
            & F/OSS package (\ref{sec:overview:maintainer}),\newline
              free infrastructure and services (\ref{sec:overview:maintainer}),\newline
              bring your own compute (\ref{sec:overview:collaborators}) \\
        \bottomrule
    \end{tabulary}
    }
\end{table}

\subsection{Design concepts}
\label{sec:conceptual:design-concepts}

To address these challenges, we think creatively about how certain data science steps might fit into a modified open-source development process. Our starting point is to look for processes in which some important functionality can be decomposed into smaller, similarly-structured patches that can be evaluated using standardized measures. Through our experience researching and developing feature engineering pipelines and systems, as well as our review of the requirements and key characteristics of the feature engineering process, we found that we could extend and adapt the pull request model to facilitate collaborative development in data science by following a series of four corresponding design concepts (\Cref{tab:concept-mapping}), which form the basis for our framework.

\begin{enumerate}[label={D\arabic*}]

  \item \hypertarget{D1}{} \textit{Data science patches.} We identify steps of the data science process that can be broken down into many \textit{patches} --- modular source code units --- which can be developed and contributed separately in an incremental process. For example, given a feature engineering pipeline, a patch is a new feature definition to be added to the pipeline.

  \item \hypertarget{D2}{} \textit{Data science products in open-source workflows.} A usable data science artifact forms a \textit{product} that is stored in an open-source repository. For example, when solving a feature engineering task, the product is an executable feature engineering pipeline. The composition of many patches from different collaborators forms a product that is stored in a repository on a source code host in which patches are proposed as individual pull requests. We design this process to accommodate collaborators of all backgrounds by providing multiple development interfaces. Notebook-based workflows are popular among data science developers, so our framework supports creation and submission of patches entirely within the notebook.

  \item \hypertarget{D3}{} \textit{Software and statistical acceptance procedures.}  ML products have the usual software quality measures along with statistical/ML performance metrics. Collaborators receive feedback on the quality of their work from both of these points of view.
    
  \item \hypertarget{D4}{} \textit{\myrevision{Decentralized development.}} A lightweight approach is needed for managing code, data, and computation. In our decentralized model, each collaborator uses their own storage and compute, and we leverage existing community infrastructure for source code management and patch acceptance.

\end{enumerate}

Besides feature engineering, how and when can this framework be used? Several conditions must be met. First, the data science product must be able to be decomposed into small, similarly-structured patches. Otherwise, the framework has a limited ability to integrate contributions. Second, human knowledge and expertise must be relevant to the generation of the data science patches. Otherwise, automation or learning alone may suffice. Third, measures of statistical and ML performance, or good proxies thereof, must be definable at the level of individual patches. Otherwise, it is difficult for maintainers to reason about how and whether to integrate patches. Finally, dataset size and evaluation time requirements must not be excessive. Otherwise, we could not use existing services that are free for open-source development.\footnote{As a rough guideline, running the evaluation procedure on the validation data should take no more than five minutes in order to facilitate interactivity.}

So while we focus on feature engineering, this framework can apply to other steps in data science pipelines --- for example, data programming with labeling functions and slicing functions \cite{ratner2016data, chen2019slicing}, which we leave for future work.

In the next section, we apply these design principles to describe a framework for collaboration on predictive modeling projects, referring back to these challenges and design concepts as they appear. Then in \Cref{sec:fteng}, we implement this general approach more specifically for collaborative feature engineering on tabular data.

\section{An overview of \Ballet}
\label{sec:overview}

\begin{figure}
  \centering
  \includegraphics[clip=true, trim={0 233 105 5}, width=\linewidth]{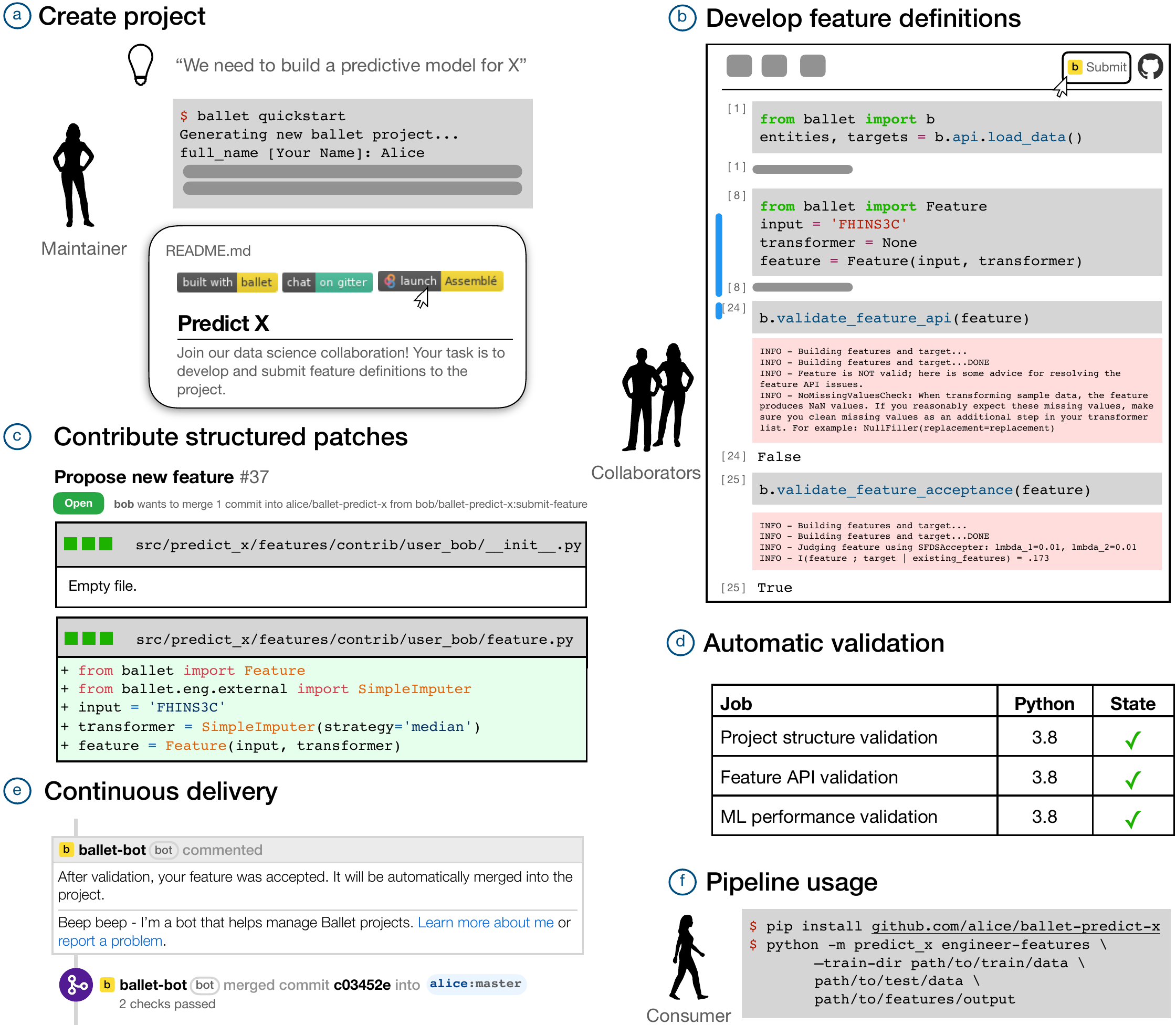}
  \caption{\myrevision{Collaborative data science development with the \Ballet framework for a feature engineering project. (a) A maintainer with a dataset wants to mobilize the power of the open data science community to solve a predictive modeling task. They use the Ballet CLI to render a new project from a provided template and push to GitHub. (b) A developer interested in the project is tasked with writing feature definitions (defining \inline{Feature} instances). They can launch the project in \Assemble, a custom development environment built on Binder and JupyterLab. Ballet's high-level client supports them in automatically detecting the project configuration, exploring the data, developing candidate feature definitions, and surfacing any API and ML performance issues. Once issues are fixed, the developer can submit the feature definition alone from within their messy notebook by selecting the code cell and using \Assemble{'s} submit button. (c) The selected code is automatically extracted and processed as a pull request following the project structure imposed by \Ballet. (d) In continuous integration, \Ballet runs feature API and ML performance validation on this one feature definition. (e) Feature definitions that pass can be automatically and safely merged by the \Ballet Bot. (f) Ballet will collect and compose this new feature definition into the existing feature engineering pipeline, which can be used by the community to modeling their own raw data (usage stylized).}}
  \label{fig:overview}
  \Description{The figure shows 6 panels. Panel A ("Create project") shows an icon of a person labeled "maintainer" thinking, "we need to build a predictive model for X". A code snippet shows them running the "ballet quickstart" command which produces output including "Generating new ballet project...". A stylized view of the resulting project's README includes three badges, "built with ballet", "chat on gitter", and "launch Assemble" and the text "Join our data science collaboration! Your task is to develop and submit feature definitions to the project." Panel B ("Develop feature definitions") shows an icon of a group labeled "collaborators". It contains a stylized version of  a computational notebook showing cell 1 which shows the Ballet client used to load raw data, cell 8 which shows a completed feature definition, cell 24 which shows the Ballet client used to validate the feature API but that the validation fails due to NoMissingValuesCheck, and cell 25 which shows the Ballet client used to validate the ML performance using the "SFDSAccepter" and succeeds because the mutual information between the feature and target conditional on the existing features is at $.173$ which must be above whatever is the threshold set for the project. At the top right of the notebook toolbar is a mouse cursor hovering over a button with the Ballet logo titled "submit". Panel C ("Contribute structured patches") shows a stylized view of an open GitHub pull request "Propose new feature (#37)" with the target "alice/ballet-predict-x" and the source "bob/ballet-predict-x:submit-feature. The corrected feature definition from Panel B is shown at "src/predict_x/features/contrib/user_bob/feature.py" along with an empty sibling file "__init__.py". Panel D ("Automatic validation") shows a table of three jobs ("Project structure validation", "Feature API validation", "ML performance validation") each associated with Python equals 3.8 and State equals a bold checkmark. Panel E ("Continuous delivery") shows a screenshot of "ballet-bot" commenting on a pull request, "After validation, your feature was accepted. it will be automatically merged into the project." Panel F ("Pipeline usage") and shows an icon of a person labeled "consumer" and a code snippet that first shows installing ballet-predict-x using pip and then using its command-line interface to apply the pipeline.}
  \vspace{-2ex}
\end{figure}

Ballet extends the open-source development process to support collaborative data science by applying the concepts of data science patches, data science products in open-source workflows, software and statistical acceptance procedures, and decentralized development. As this process is complex, we illustrate how \Ballet works by showing the experience of using it from three perspectives --- maintainer, collaborator, and consumer --- building on existing work that investigates different users' roles in open source development and ecosystems~\cite{yu2007mining, berdou2010organization, roberts2006understanding, barcellini2014situated, hauge2010adoption}. This development cycle is illustrated in \Cref{fig:overview}. In \Cref{sec:fteng}, we present a more concrete example of feature engineering on tabular datasets.

\subsection{Maintainer}
\label{sec:overview:maintainer}

A maintainer wants to build a predictive model. They first define the prediction goal and upload their dataset. They install the \Ballet package, which includes the core framework libraries and command line interface (CLI). Next, they use the CLI to automatically render a new repository from the provided project template, which contains the minimal files and structure required for their project, such as directory organization, configuration files, and problem metadata. \myrevision{They define a task for collaborators: create and submit a data science patch that performs well (\hyperlink{C1}{C1}/\hyperlink{D1}{D1}) --- for example, a feature definition that has high predictive power. The resulting repository contains a usable (if, at first, empty) data science pipeline (\hyperlink{C2}{C2}/\hyperlink{D2}{D2}).} After pushing to GitHub and enabling our CI tools and bots, the maintainer begins recruiting collaborators.

Collaborators working in parallel submit patches as pull requests with a careful structure provided by \Ballet. Not every patch is worthy of inclusion in the product. As patches arrive from collaborators in the form of pull requests, the CI service is repurposed to run \Ballet's acceptance procedure such that only high-quality patches are accepted (\hyperlink{C3}{C3}/\hyperlink{D3}{D3}). This ``working pipeline invariant'' aligns data science pipelines with the aim of continuous delivery in software development \cite{humble2010continuous}. In feature engineering, the acceptance procedure is a feature validation suite (\Cref{sec:fteng:testing}) which marks individual feature definitions as accepted/rejected, and the resulting feature engineering pipeline on the default branch can always be executed to engineer feature values from new data instances.

One challenge for maintainers is to integrate data science patches as they begin to stream in. Unlike software projects where contributions can take any form, these types of patches are all structured similarly, and if they validate successfully, they may be safely merged without further review. To support maintainers, the \textit{\Ballet Bot}\footnote{\url{\balletboturl}} can automatically manage contributions, performing tasks such as merging pull requests of accepted patches and closing rejected ones. The process continues until either the performance of the ML product exceeds some threshold, or improvements are exhausted.

\Ballet projects are lightweight, as our framework is distributed as a free and open-source Python package, and use only lightweight infrastructure that is freely available to open-source projects, like GitHub, Travis, and Binder (\hyperlink{C4}{C4}/\hyperlink{D4}{D4}). This avoids spinning up data stores or servers --- or relying on large commercial sponsors to do the same.

\subsection{\myrevision{Collaborators}}
\label{sec:overview:collaborators}

A data science developer is interested in the project and wants to contribute. They find the task description and begin learning about the project and about \Ballet. They can review and learn from existing patches contributed by others and discuss ideas in an integrated chatroom. They begin developing a new patch in their preferred development environment (\emph{patch development task}). When they are satisfied with its performance, they propose to add it to the upstream project at a specific location in the directory structure using the pull request model (\emph{patch contribution task}). In designing \Ballet, we aimed to make it as easy as possible for data science developers with varying backgrounds to accomplish the patch development and patch contribution tasks within open-source workflows.

The \Ballet interactive client, included in the core library, supports developers in loading data, exploring and analyzing existing patches, validating the performance of their work, and accessing functionality provided by the shared project. It also decreases the time to beginning development by automatically detecting and loading a Ballet project's configuration. Use of the client is shown in \Cref{fig:overview}, Panel B.

We enable several interfaces for collaborators to develop and submit patches like feature definitions, where different interfaces are appropriate for collaborators with different types of development expertise, relaxing requirements on the development style of collaborators. \myrevision{Collaborators} who are experienced in open-source development processes can use their preferred tools and workflow to submit their patch as a pull request, supported by the \Ballet CLI --- the project is still just a layer built on top of familiar technologies like git. However, in preliminary user studies (\Cref{sec:user-studies:preliminary-studies}), we found that adapting from usual data science workflows was a huge obstacle. Many data science developers we worked with had never successfully used open-source development processes to contribute to any shared project.

We addressed this by centering all development in the notebook with \emph{\Assemble},\footnote{\url{\assembleurl}} a cloud-based workflow and development environment for contribution to \Ballet projects (\hyperlink{C1}{C1}/\hyperlink{D1}{D1}). We created a custom experience on top of community tooling that enables data science developers to develop and submit features entirely within the notebook.

\Assemble consists of several pieces. First, we support the patch development task by easing the process for developers to set up a computational environment and begin exploratory analysis.
We tightly integrate \Assemble to be deployed on Binder,\footnote{\url{https://mybinder.org/}} a free community service for cloud-hosted notebooks, such that \Assemble can be launched from every \Ballet project from a hyperlinked ``badge.'' Following a step-by-step guide, developers create a patch within a messy, exploratory notebook. Second, we support the patch contribution task through a new, entirely in-notebook interface. We design and implement a JupyterLab extension for submitting code to a \Ballet project using a simple, one-click ``Submit'' button.  Developers isolate a patch, such as a single feature definition, in a code cell and click to submit it. This causes the patch to be first preprocessed and subjected to initial server-side validation before being formulated as a pull request on the \myrevision{collaborator}'s behalf. Low-level version control details are abstracted away from users, and \myrevision{collaborators} can view their new pull request with its validation results in a matter of seconds.

The submitted feature definitions are marked as accepted or rejected, and \myrevision{collaborator}s can proceed accordingly, either moving on to their next idea or reviewing diagnostic information and trying to fix a rejected submission.

\subsection{Consumer}
\label{sec:overview:consumer}

\Ballet project consumers --- members of the wider data science community --- are now free to use the data science product however they desire, such as by executing the feature engineering pipeline to extract features from new raw data instances, making predictions for their own datasets. The project can easily be installed using a package manager like \inline{pip} as a versioned dependency, and ML engineers can extract the machine-readable feature definitions into a feature store or other environment for further analysis and deployment. This ``pipeline as code''/``pipeline as package'' approach makes it easy to use and re-use the pipeline and helps address data lineage/versioning issues.

For example, the \Ballet client for feature engineering projects exposes a method that fits the feature engineering pipeline on training data and can then engineer features from new data instances, and an instance of a pipeline that is already fitted on data from of the upstream project. These can be easily used in other library code.

\section{Collaborative feature engineering}
\label{sec:fteng}

We now \myrevision{briefly} describe the design and implementation of a feature engineering plugin for \Ballet for collaborative feature engineering projects. While we spoke in general terms about data science patches and products and the statistical acceptance procedure, here we define these concepts for feature engineering. \myrevision{Full details are presented in \Cref{app:fteng}.}

We start from the insight that feature engineering can be represented as a dataflow graph over individual features. We structure code that extracts a group of feature values as a patch, calling these \textit{feature definitions} and representing them with a \inline{Feature} interface. Feature definitions are composed into a feature engineering pipeline product. Newly contributed feature definitions are accepted if they pass a two-stage acceptance procedure that tests both the feature API and its contribution to ML performance. Finally, the plugin specifies a module organization that allows features to be collected programmatically.

In Ballet, we create a flexible and powerful language for feature engineering that is embedded within the larger framework. It supports functionality such as learned feature transformations, supervised feature transformations, nested transformer steps, syntactic sugar for functional transformations, data frame-style transformations, and recovery from errors due to type incompatibility.

\begin{figure}[t]
  \begin{subfigure}[T]{0.46\linewidth}
\begin{minted}{./listings/customlexer.py:PythonLexer -x}
from ballet import Feature
from ballet.eng import ConditionalTransformer
from ballet.eng.external import SimpleImputer
import numpy as np

input = 'Lot Area'
transformer = [
    ConditionalTransformer(
        lambda ser: ser.skew() > 0.75,
        lambda ser: np.log1p(ser)),
    SimpleImputer(strategy='mean'),
]
name = 'Lot area unskewed'
feature = Feature(input, transformer, name=name)
\end{minted}
    \caption{A feature definition that conditionally unskews lot area (for a house price prediction problem) by applying a log transformation only if skew is present in the training data and then mean-imputing missing values.}
    \label{fig:feature-examples:ames}
  \end{subfigure}%
  \quad
  \begin{subfigure}[T]{0.51\linewidth}
\begin{minted}{./listings/customlexer.py:PythonLexer -x}
from ballet import Feature
from ballet.eng.external import SimpleImputer
import numpy as np

input = 'JWAP'  # Time of arrival at work
transformer = [
    SimpleImputer(missing_values=np.nan, strategy='constant', fill_value=0.0),
    lambda df: np.where((df >= 70) & (df <= 124), 1, 0),
]
name = 'If job has a morning start time'
description = 'Return 1 if the Work arrival time >=6:30AM and <=10:30AM'
feature = Feature(input, transformer, name=name, description=description)
\end{minted}
    \caption{A feature definition that defines a transformation of work arrival time (for a personal income prediction problem) by filling missing values and then applying a custom function.}
    \label{fig:feature-examples:census}
  \end{subfigure}

  \caption{Examples of user-submitted feature definitions in two different \Ballet projects. \Featurefunctions are collected and imported from standalone modules like these and composed into a single feature engineering pipeline that separately applies each \featurefunction and concatenates resulting feature values.}
  \label{fig:feature-examples}
  \Description{Two code listings of Python code.}
\end{figure}

\subsection{Feature definitions}
\label{sec:fteng:feature-definitions}

A \textit{feature definition} is the code that is used to extract semantically related feature values from raw data. Let us observe data instances ${\D = (\mathbf{v}_i, \mathbf{y}_i)_{i=1}^{N}}$, where $\mathbf{v}_i \in \V$ are the raw variables and $\mathbf{y}_i \in \Y$ is the target. In this formulation, the raw variable domain $\V$ includes strings, missing values, categories, and other non-numeric types that cannot typically be inputted to learning algorithms. Thus our goal in feature engineering is to develop a learned map from $\V$ to $\X$ where $\X \subseteq \R^n$ is a real-valued feature space.

\begin{definition}
A \emph{\featurefunction} is a learned map from raw variables in one data instance to feature values, $f: (\V, \Y) \to \V \to \X$.
\end{definition}

Each \featurefunction learns a specific map from $\D$, such that any information it uses, such as variable means and variances, is learned from the development (training) dataset. This formalizes the separation between development and testing data to avoid any \emph{leakage} of information during the feature engineering process.

The \inline{Feature} class in Ballet is a \myrevision{way to express a \featurefunction in code}. It is a tuple (\inline{input}, \inline{transformer}). The input declares the variable(s) from $\V$ that are needed by the feature, which will be passed to \inline{transformer}, one or more transformer steps. Each transformer step implements the learned map via \inline{fit} and \inline{transform} methods, a standard interface in machine learning pipelines \cite{buitinck2013api}. A data science developer then simply provides values for the input and transformer of a \inline{Feature} object in their code. \myrevision{Additional metadata, like \inline{name}, \inline{description}, and \inline{source}, can also be provided.} Two example feature definitions are shown in \Cref{fig:feature-examples}.

\subsection{Feature engineering pipelines}
\label{sec:fteng:feature-engineering-pipelines}

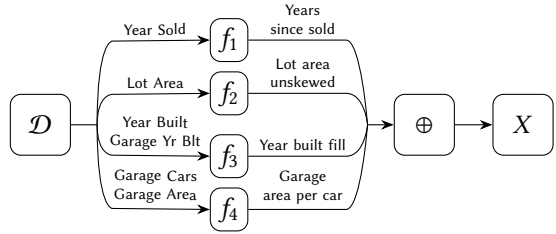
\begin{figure}[ht]
  \centering
  \begin{tikzpicture}[
    sourcesink/.style={rectangle, rounded corners, text centered, draw=black, minimum width=0.8cm, minimum height=0.8cm},
  feature/.style={rectangle, rounded corners, text centered, draw=black, minimum height=0.5cm},
  arrow/.style={->,>=Stealth},
  arrowtextleft/.style={xshift=-0.5ex,yshift=2pt,text centered, text width=1.3cm, inner sep=0pt, font=\sffamily\tiny,sloped},
  arrowtextright/.style={xshift=0.5ex,yshift=2pt,text centered, text width=1.3cm, inner sep=0pt, font=\sffamily\tiny,sloped},
    ]

  \node (D) at (0,0) [sourcesink] {$\mathcal{D}$};
  \node (aux0) [right=0.1cm of D, minimum width=0.0cm] {};
  \node (f1) [feature, above right=0.7cm and 1.5cm of aux0] {$f_1$};
  \node (f2) [feature, above right=0.0cm and 1.5cm of aux0] {$f_2$};
  \node (f3) [feature, below right=0.0cm and 1.5cm of aux0] {$f_3$};
  \node (f4) [feature, below right=0.7cm and 1.5cm of aux0] {$f_4$};
  \node (aux1) [right=3.6cm of aux0, minimum width=0.0cm] {};
  \node (oplus) [sourcesink, right=0.1cm of aux1] {$\oplus$};
  \node (X) [sourcesink, right=0.5cm of oplus] {$X$};

  \draw (D.east) -- (aux0.east);
  \draw [arrow] (aux0.east) .. controls +(0.0,1.1) and +(0.0,0) .. +(0.2,1.1) .. controls +(0,0) and +(0,0) .. node[arrowtextleft,above] {Year Sold} (f1.west);
   \draw [arrow] (aux0.east) .. controls +(0.0,0.4) and +(0.0,0) .. +(0.2,0.4) .. controls +(0,0) and +(0,0) .. node[arrowtextleft,above] {Lot Area} (f2.west);
   \draw [arrow] (aux0.east) .. controls +(0.0,-0.4) and +(0.0,0) .. +(0.2,-0.4) .. controls +(0,0) and +(0,0) .. node[arrowtextleft,above] {Year Built\\Garage Yr Blt} (f3.west);
   \draw [arrow] (aux0.east) .. controls +(0.0,-1.1) and +(0.0,0) .. +(0.2,-1.1) .. controls +(0,0) and +(0,0) .. node[arrowtextleft,above] {Garage Cars\\Garage Area} (f4.west);
   \draw [arrow] (aux1.west) -- (oplus.west);
   \draw (f1.east) .. node[arrowtextright,above] {Years since sold} controls +(0.0,0) and +(0.0,0) .. +(1.3,0) .. controls +(0.2,0) and +(0,0) .. (aux1.west);
   \draw (f2.east) .. node[arrowtextright,above] {Lot area unskewed} controls +(0.0,0) and +(0.0,0) .. +(1.3,0) .. controls +(0.25,0) and +(0,0) .. (aux1.west);
   \draw (f3.east) .. node[arrowtextright,above] {Year built fill} controls +(0.0,0) and +(0.0,0) .. +(1.3,0) .. controls +(0.25,0) and +(0,0) .. (aux1.west);
   \draw (f4.east) .. node[arrowtextright,above] {Garage area per car} controls +(0.0,0) and +(0.0,0) .. +(1.3,0) .. controls +(0.2,0) and +(0,0) .. (aux1.west);
   \draw [arrow] (oplus) -- (X);

\end{tikzpicture}
  \caption{A feature engineering pipeline for a house price prediction problem with four \featurefunctions operating on six raw variables.}
  \label{fig:feature-engineering-pipeline-example}
  \Description{A directed graph with a source node labeled D, four intermediate nodes labeled f1, f2, f3, and f4, another node labeled with the concatenation operator, and the sink node labeled X. The edge from D to f1 is labeled "Year Sold," the edge from D to f2 is labeled "Lot Area," the edge from D to f3 is labeled "Year Built/Garage Yr Blt" and the edge from D to f4 is labeled "Garage Cars/Garage Area." The edge from f1 to concatenation is labeled "Years since sold," the edge from f2 to concatenation is labeled "Lot area unskewed," the edge from f3 to concatenation is labeled "Year built fill," and the edge from f4 to concatenation is labeled "Garage area per car."}
\end{figure}

Features are then composed together in a feature engineering pipeline. A \emph{feature engineering pipeline} applies each \featurefunction to the new data instances and concatenates the result, yielding a feature matrix $X$. This is implemented in \Ballet by the \inline{FeatureEngineeringPipeline} class that operates directly on data frames. Each \featurefunction within the pipeline is passed the input variables it requires that it then transforms appropriately, internally using a sequence of one or more transformer steps (\Cref{fig:feature-engineering-pipeline-example}).

\subsection{Acceptance procedures for feature definitions}
\label{sec:fteng:testing}

Contributions of feature engineering code, just like other code contributions, must be evaluated for quality before being accepted, at risk of introducing errors, malicious behavior, or design flaws. For example, a \featurefunction that produces non-numeric values can result in an unusable feature engineering pipeline. Large feature engineering collaborations can also be susceptible to ``feature spam,'' a high volume of low-quality feature definitions (submitted either intentionally or accidentally) that harm the collaboration \cite{smith2017featurehub}. Modeling performance can suffer and require an additional feature selection step --- violating the working pipeline invariant --- and the experience of other collaborators can be harmed who are not able to assume that existing feature definitions are high-quality.

To address these possibilities, we extensively validate feature definition contributions for software quality and ML performance, implemented as a test suite that is both exposed by the \Ballet client and executed in CI for every pull request. Thus the same method that is used in CI for validating feature contributions is available to data science developers for debugging and performance evaluation in their development environment. \Ballet Bot can automatically merge pull requests corresponding to accepted feature definitions and close pull requests corresponding to rejected feature definitions.

\paragraph{Feature API validation}
\label{sec:fteng:testing:feature-testing}

User-contributed feature definitions should satisfy the \inline{Feature} interface and successfully deal with common error situations, such as intermediate computations producing missing values. We fit the \featurefunction to a separate subsampled training dataset in an isolated environment and extract feature values from subsampled training and validation datasets, failing immediately on any implementation errors. We then conduct a battery of 15 tests to increase confidence that the \featurefunction would also extract acceptable feature values on unseen inputs (\Cref{tab:test-suite-feature-api-validation}). Each test is paired with ``advice'' that can be surfaced back to the user to fix any issues (\Cref{fig:overview}).

Another part of feature API validation is an analysis of the changes introduced in a proposed PR to ensure that the required project structure is preserved and that the collaborator has not accidentally included irrelevant code that would need to be evaluated separately.

\paragraph{ML performance validation}
\label{sec:fteng:testing:ml-performance-validation}

A complementary aspect of the acceptance procedure is validating a feature contribution in terms of its impact on machine learning performance, which we cast as a streaming feature definition selection (SFDS) problem. This is a variant of streaming feature selection where we select from among feature definitions rather than feature values. Features that improve ML performance will pass this step; otherwise, the contribution will be rejected. Not only does this discourage low-quality contributions, but it provides a way for collaborators to evaluate their performance, incentivizing more deliberate and creative feature engineering.

We first compile requirements for an SFDS algorithm to be deployed in our setting, including that the algorithm should be stateless, support real-world data types (mixed discrete and continuous), and be robust to over-submission. While there is a wealth of research into streaming feature selection \cite{zhou2005streaming, wu2013online, wang2015online, yu2016scalable}, no existing algorithm satisfies all requirements. Instead, we extend prior work for our situation. Our SFDS algorithm proceeds in two stages (\Cref{app:fteng:testing:streaming-feature-definition-selection}). In the \textit{acceptance} stage, we compute the conditional mutual information of the new feature values with the target given the existing feature matrix and accept the feature if it is above a dynamic threshold. In the \textit{pruning} stage, existing features that have been made newly redundant by accepted features can be pruned. \myrevision{Maintainers of individual projects are also free to configure alternative ML performance validation algorithms given their needs, and \Ballet provides several such implementations.}

\section{User studies}
\label{sec:user-studies}

The success of a collaborative framework must be evaluated in the context of its usage. To that extent, we report on three user studies, including a large-scale case study.

\subsection{Preliminary studies}
\label{sec:user-studies:preliminary-studies}

We conducted several preliminary studies and evaluation steps during the iterative design process for \Ballet.

\paragraph{Disease incidence prediction}

We first evaluated an initial prototype of \Ballet in a user study with eight data science developers. All participants had at least basic knowledge of collaborative software engineering and open-source development practices (i.e., using git and pull requests). We explained the framework and gave a brief tutorial on how to write feature definitions. Participants were then tasked with writing feature definitions to help predict the incidence of dengue fever given historical data from Iquitos, Peru and San Juan, Puerto Rico \cite{dengueforecasting}, for which they were allotted 30 minutes.
Three participants were successfully able to merge their first feature definition within this period, while the remainder produced features with errors or were unable to write one. In interviews, participants suggested that they found the \Ballet framework helpful for structuring contributions and validating features, but were unfamiliar with writing feature engineering code in terms of feature definitions with separate fit/transform behavior (\Cref{sec:fteng:feature-definitions}), and struggled to translate exploratory work in notebooks all the way to pull requests to a shared project. Based on this feedback, we created the \inline{ballet.eng} library of feature engineering primitives (\Cref{app:fteng:feature-engineering-primitives}) and created tutorial materials for new \myrevision{collaborator}s. We also began the design process that became the \Assemble environment that supports notebook-based developers (\Cref{sec:overview:collaborators}).

\paragraph{House price prediction}

We evaluated a subsequent version of \Ballet in a user study with 13 researchers and data science developers. This version included changes made since the first preliminary study and introduced an alpha version of \Assemble that did not yet include server-side or in-notebook validation functionality. Five of the participants had little to no prior experience contributing to open-source projects, six reported contributing occasionally, and two contributed frequently. All self-reported as intermediate or expert data scientists and Python developers. Participants were given a starter notebook that guided the development and contribution of feature definitions, and documentation on the framework. They were tasked with writing feature definitions to help predict the selling price of a house given administrative data collected in Ames, Iowa \cite{decock2011ames}. After contributing, participants completed a short survey with a usability evaluation and provided semi-structured free-text feedback. Participants reported that they were moderately successful at learning to write and submit feature definitions but wanted more code examples. They also reported that they wanted to validate their features within their notebook using the same methods that were used in the automated testing in CI. Based on this feedback, among other improvements, we expanded and improved our feature engineering guide and the starter notebook. We also made methods for feature API validation and ML performance validation available in the interactive client (\Cref{sec:fteng:testing}) and expanded the \Assemble server-side validation to catch common issues.

\subsection{Case study}
\label{sec:user-studies:case-study}

Finally, we conducted a full user study with the versions of \Ballet and \Assemble described in this paper. To better understand the characteristics of live collaborative data science projects, we use a mixed-method software engineering case study approach~\cite{runeson2009guidelines}. The case study approach allows us to study the phenomenon of collaborative data science in its ``real-life context.'' This choice of evaluation methodology allows us to move beyond a laboratory setting and gain deeper insights into how large-scale collaborations function and perform. Through this study, we aim to answer the following four research questions:

\begin{enumerate}[label={\textbf{RQ\arabic*}}]
  \item What are the most important aspects of our collaborative framework to support participant experience and project outcomes?
  \item What is the relationship between participant background and participant experience/performance?
  \item What are the characteristics of feature engineering code in a collaborative project?
  \item How does a collaborative model perform in comparison to other approaches?
\end{enumerate}

These research questions build on our conceptual framework, allowing us to better understand the effects of our design choices as well as move to forward our understanding of collaborative data science projects in general.

\paragraph{General procedures}

We created an open-source project using \Ballet, \pci{,}\footnote{\url{\balletcensusurl}} to produce a feature engineering pipeline for personal income prediction. After invited participants consented to the research study terms, we asked them to fill out a pre-participation survey with background information about themselves, which served as the independent variables of our study.
Next, they were directed to the public repository containing the collaborative project and asked to complete the task described in the project README. They were instructed to use either their preferred development environment or \Assemble. After they completed this task, we surveyed them about their experience.

\paragraph{Participants}

In recruiting participants, we wanted to ensure that our study included beginners and experts in data science, software development, and survey data analysis (the problem domain). To achieve this, we compiled personal contacts with various backgrounds. After reaching these contacts, we then used snowball sampling to recruit more participants with similar backgrounds. We expanded our outreach by posting to relevant forums and mailing lists in data science development, Python programming, and survey data analysis. Participants were entered into a drawing for several nominal prizes but were not otherwise compensated.

\paragraph{Dataset}

The input data is the raw survey responses to the 2018 U.S. Census American Community Survey (ACS) for Massachusetts (\Cref{tab:acs-description}). This ``Public Use Microdata Sample'' (PUMS) has anonymized individual-level responses. Unlike the classic ML ``\myrevision{adult} census'' dataset \cite{kohavi1996scaling} which is highly preprocessed, raw ACS responses are a realistic form for a dataset used in an open data science project. Following \citet{kohavi1996scaling}, we define the prediction target as whether an individual respondent will earn more than \$84,770 in 2018 (adjusting the original prediction target of \$50,000 for inflation), and filter a set of ``reasonable'' rows by keeping people older than 16 with personal income greater than \$100 with hours worked in a typical week greater than zero. We merged the ``household'' and ``person'' parts of the survey to get compound records and split the survey responses into a development set and a held-out test set. 
\begin{table}[ht]

  \caption{ACS dataset used in \pci project.}
  \label{tab:acs-description}

  \centering
    \begin{tabular}{lcc}
    \toprule
    {} &  Development &   Test \\
    \midrule
    Number of rows &  30085 &  10029 \\
    Entity variables &    494 &    494 \\
      High income    &   7532 &   2521 \\
    Low income     &  22553 &   7508 \\
    \bottomrule
  \end{tabular}

\end{table}

\paragraph{\myrevision{Research} instruments}

Our mixed-method study synthesizes and triangulates data from five sources:

\begin{itemize}[labelindent=\parindent]

  \item \textit{Pre-participation survey.} Participants provided background information about themselves, such as their education; occupation; self-reported background with data science, feature engineering, Python programming, open-source development, analysis of survey data, and familiarity with the U.S. Census/ACS specifically; and preferred development environment. Participants were also asked to opt in to telemetry data collection.

  \item \textit{\Assemble telemetry.} To better understand the experience of participants who use \Assemble on Binder, we instrumented the extension and installed an instrumented version of \Ballet to collect usage statistics and some intermediate outputs.
    Once participants authenticated with GitHub, we checked with our telemetry server to see whether they had opted in to telemetry data collection. If they did so, we sent and recorded the buffered telemetry events.

  \item \textit{Post-participation survey.} Participants who attempted and/or completed the task were asked to fill out a survey about their experience, including the development environment they used, how much time they spent on each sub-task, and which activities they did and functionality they used as part of the task and which of these were most important. They were also asked to provide open-ended feedback on different aspects, and to report how demanding the task was using the NASA-TLX Task Load Index \citep{hard1988nasatlx}, a workload assessment that is widely used in usability evaluations in software engineering and other domains \citep{cook2005user, salman2018effect}. Participants indicate on a scale the temporal demand, mental demand, and effort required by the task, their perceived performance, and their frustration. The TLX score is a weighted average of responses (0=very low task demand, 100=very high task demand).

  \item \textit{Code contributions.} For participants who progressed in the task to the point of submitting a feature definition to the upstream \pci project, we analyze both the submitted source code as well as the performance characteristics of the submission.

  \item \textit{Expert and AutoML baselines.} To obtain comparisons to solutions born from Ballet collaborations, we also obtain baseline solutions to the personal income prediction problem from outside data science experts and from a cloud provider's AutoML service. First, we asked two outside data science experts working independently to solve the combined feature engineering and modeling task (without knowledge of the collaborative project).\footnote{Replication files are available at \url{\replicationurl}.} These experts were asked to work until they were satisfied with the performance of their predictive model, but not to exceed four hours, and were not compensated. Second, we used Google Cloud AutoML Tables,\footnote{\url{https://cloud.google.com/automl-tables/}} an AutoML service for tabular data, which supports structured data ``as found in the wild,'' to automatically solve the task, and ran it with its default settings until convergence.

\end{itemize}

We include the study description, pre-participation survey, and post-participation survey in our supplementary materials.

\paragraph{Analysis}

After linking our data sources together, we performed a quantitative analysis to summarize results (e.g., participant backgrounds, average time spent) and relate measures to each other (e.g., participant expertise to cognitive load).
Where appropriate, we also conducted statistical tests to report on significant differences for phenomena of interest.
For qualitative analysis, we employed open and axial coding methodology to categorize the free-text responses and relate codes to each other to form emergent themes~\cite{bohm2004theoretical}.
Two researchers first coded each response independently, and responses could receive multiple codes, which were then collaboratively discussed.
We resolved disagreements by revisiting the responses, potentially introducing new codes in relation to themes discovered in other responses.
We later revisited all responses and codes to investigate how they relate to each other, which led us to the emergent themes we present in our results.
Finally, to understand the kind of source code that is produced in a collaborative data science setting, we performed
lightweight program analysis to extract and quantify the feature engineering primitives used by our participants.

\section{Results}
\label{sec:results}

We present our results by interleaving the outcomes of quantitative and qualitative analysis (including verbatim quotes from free-text responses) to form a coherent narrative around our research questions.

In total, 50 people signed up to participate in the case study and 27 people from four global regions completed the task in its entirety. To the best of our knowledge, this makes our project the sixth largest ML modeling collaboration hosted on GitHub in terms of code contributors (\Cref{tab:project-sizes}). During the case study, 28 features were merged that together extract 32 feature values from the raw data. Of case study participants, 26 submitted at least one feature and 22 had at least one feature merged.
As we went through participants' qualitative feedback about their experience, several key themes emerged, which we discuss inline.

\subsection{RQ1: Collaborative framework design}
\label{sec:results:collaborative-framework-design}

We identified several themes that relate to the design of frameworks for collaborative data science. We start by connecting these themes to the design decisions we made about \Ballet.

\textbf{Goal Clarity.} The project-level goal is clear --- to produce a predictive model. In the case of survey data that requires feature engineering, \Ballet takes the approach of decomposing this data into \myrevision{individual} goals via the feature definition abstraction, and asking collaborators to create and submit a patch that introduces a well-performing feature. Success in this task is validated using statistical tests (\Cref{sec:fteng:testing}). However, the relationship between the \myrevision{individual} and project goals may not always appear aligned to all participants. This negatively impacted some participants' experiences by introducing confusion about the direction and goal of their task. Some of the concerns expressed had to do with specific documentation elements, but others indicated a deeper confusion: \textit{``Do the resulting features have to be `meaningful' for a human or can they be built as combinations that maximize some statistical measure?''} (P2). Using the concept of software and statistical acceptance procedures, many high-quality features were merged into the project. However, the procedure was not fully transparent to the case study participants and may have prevented them from optimizing their features. While a feature that maximizes some statistical measure is best in the short term, it may constrain group productivity overall, as other participants benefit from being able to learn from existing features. And while having specific \myrevision{individual} goals incentivizes high-quality feature engineering, participants are then less focused on the project-level goal and maintainers must either implement new project functionality themselves or define additional individual goals. This is a classic tension in designing collaborative mechanisms when it comes to appropriately structuring goals and incentives~\cite{ouchi1979conceptual}.

\begin{figure}
  \centering
  \includegraphics[width=0.8\linewidth]{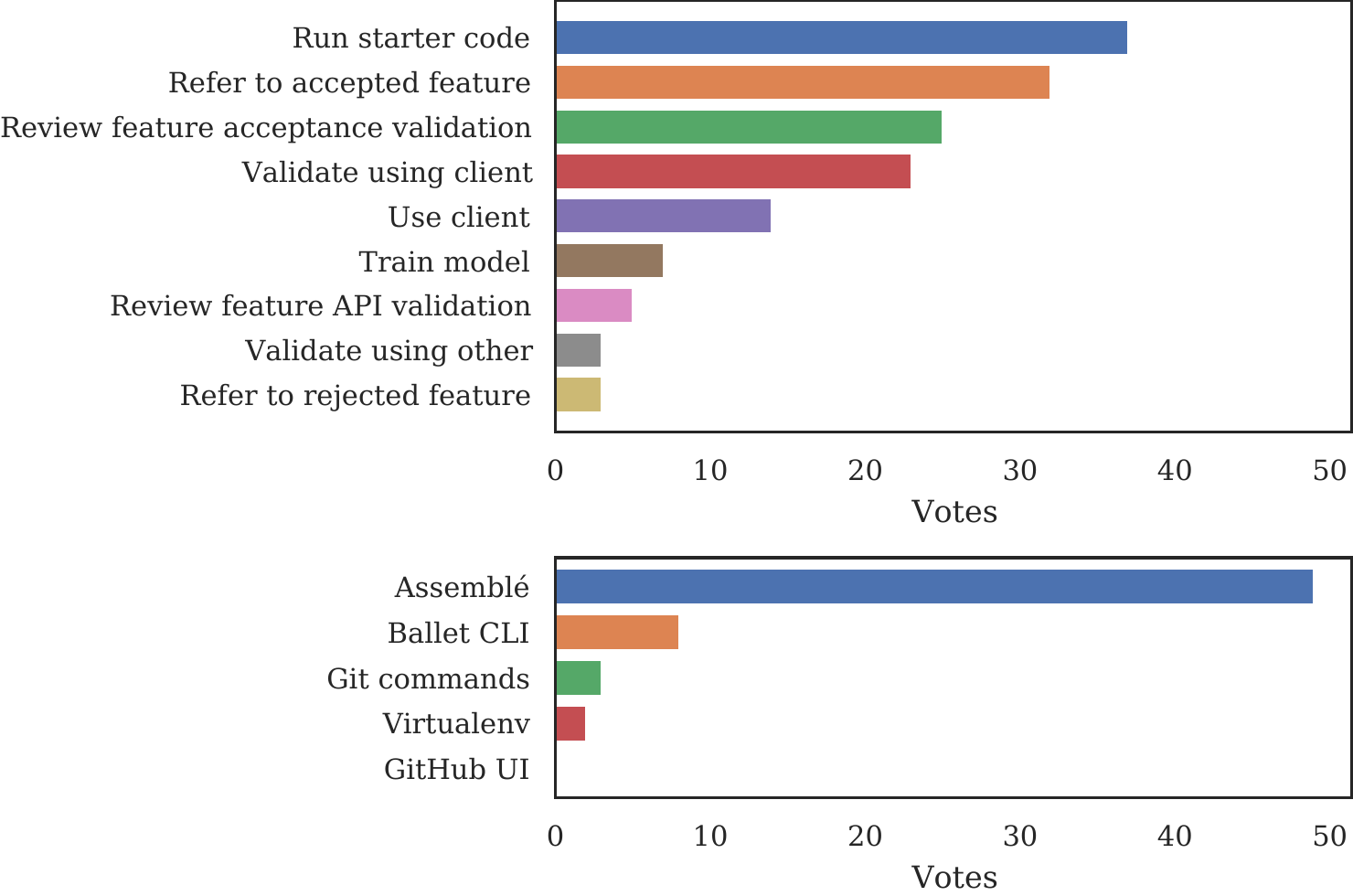}
  \caption{Most important functionality within a collaborative feature engineering project, for the patch development task (top) and the patch contribution task (bottom), according to participant votes. Participants were asked to rank their top three items for creating feature definitions (awarded three, two, and one points in aggregating votes) and their top two items for submitting feature definitions (awarded two and one points in aggregating votes).}
  \label{fig:most-important-functionality}
  \Description{The figure shows two bar charts with the dependent axis labeled "votes". The bars are as follows with approximate votes in parentheses. On the top bar chart: "Run starter code" (40), "Refer to accepted feature" (35), "Review feature acceptance validation" (25), "Validate using client" (25), "Use client" (15), "Train model" (10), "Review feature API validation" (5), "Validate using other" (3), "Refer to rejected feature" (3). On the bottom bar chart: "Assemble" (50), "Ballet CLI" (10), "Git commands" (3), "Virtualenv" (2), "GitHub UI" (0).}
\end{figure}

\textbf{Learning by Example.} We asked participants to rank the functionalities that were most important for completing the task, focusing both on creating and submitting feature definitions (\Cref{fig:most-important-functionality}). For the patch development task, participants ranked most highly the ability to refer to example code written by fellow participants or project maintainers. This form of implicit collaboration was useful for participants to accelerate the onboarding process, learn new feature engineering techniques, and coordinate their efforts.

\textbf{Distribution of Work.} However, this led to feedback about difficulties in identifying how to effectively participate in the collaboration. Participants wanted the framework to provide more functionality to determine how to partition the input space: \textit{``for better collaboration, different users can get different subsets of variables''} (P1). Some participants specifically asked for methods to review the input variables that had and had not been used and to limit the number of variables that one person would need to consider. This is a promising direction for future work, and similar ideas appear in automatic code reviewer recommendation \cite{peng2018mentionbot}. Other participants, however, were satisfied with a more passive approach in which they used the \Ballet client to programmatically explore existing feature definitions.

\textbf{Cloud-Based Workflow.} In terms of the patch contribution task, the most popular element by far was \Assemble.
Importantly, all of the nine participants who reported that they ``never'' contribute to open-source software were able to successfully submit a PR to the \pci project with \Assemble --- seven in the cloud and the others locally.\footnote{Local use involves installing JupyterLab and \Assemble on a local machine, rather than using the version running on Binder.} Attracting participants like these who are not experienced software developers is critical for sustaining large collaborations, and prioritizing interfaces that provide first-class support for collaboration can support these developers. The adaptation of the open-source development process reflected in \Assemble shows that concepts of open-source workflows and decentralized development did effectively address the aforementioned challenges for some developers.

In summary, we found that the feature definition abstraction, the cloud-based workflow in \Assemble, and the coordination and learning from referring to shared feature definitions were the aspects that contributed most to the participants' experiences. While the concept of data science patches makes significant progress toward addressing task management challenges, frictions remain around goal clarity and division of work, which should be addressed in future designs.

\subsection{RQ2: Participant background, experience, and performance}
\label{sec:results:participant-experience}

\begin{figure}
  \centering
  \includegraphics[width=0.8\linewidth]{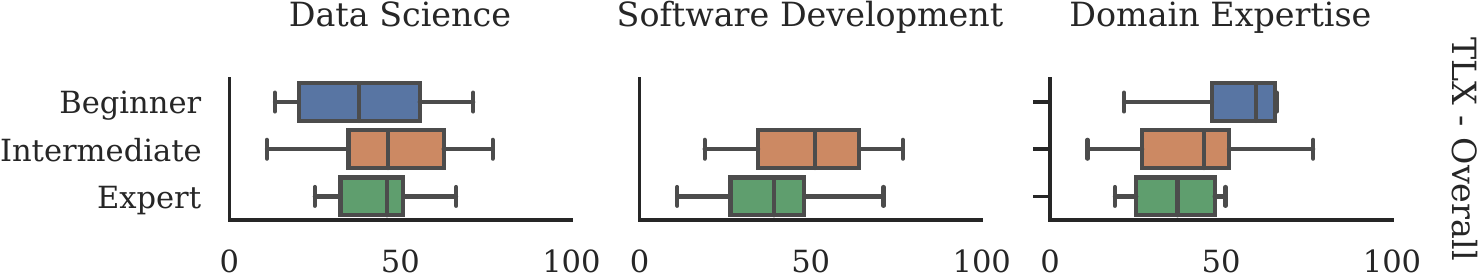}
  \\ \vspace{1ex}
  \includegraphics[width=0.8\linewidth]{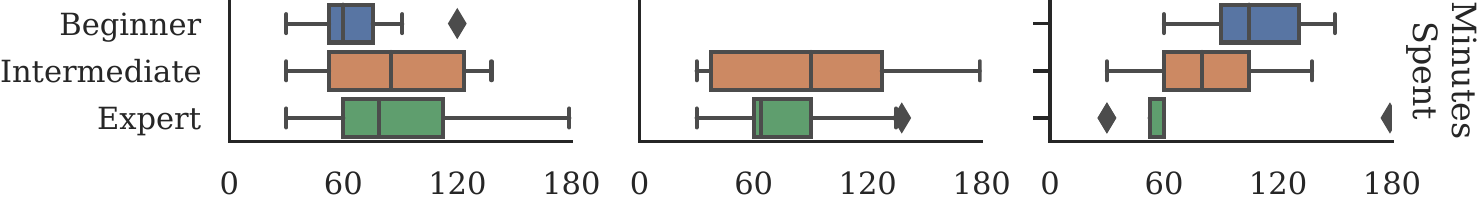}
  \\ \vspace{1ex}
  \includegraphics[width=0.8\linewidth]{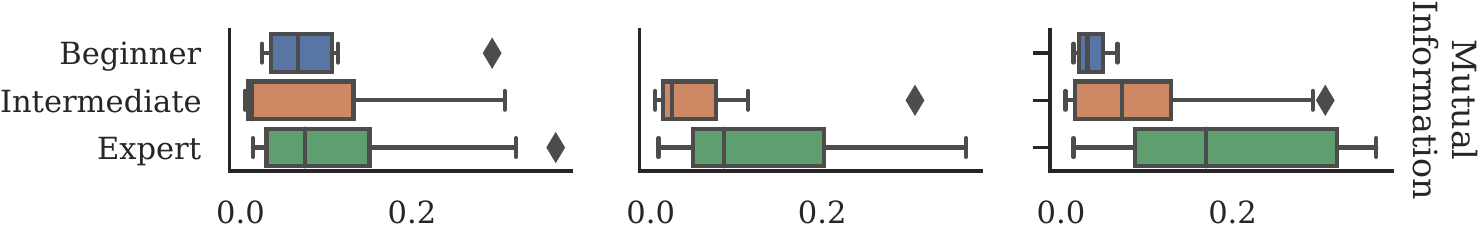}
  \\ \vspace{1ex}
  \includegraphics[width=0.8\linewidth]{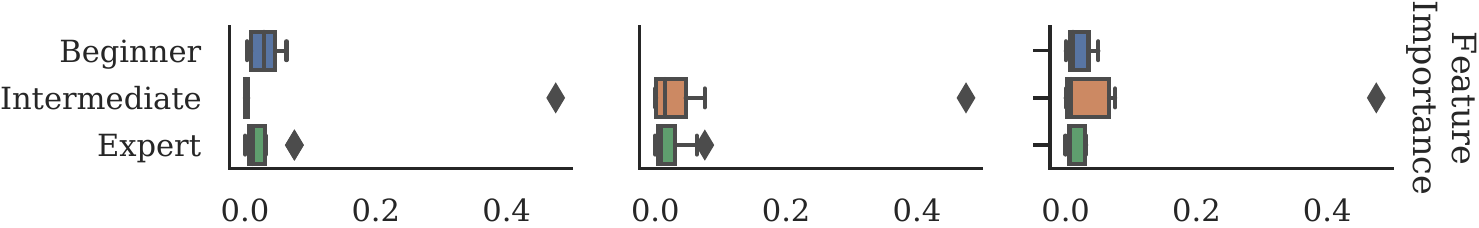}
  \caption{Task demand, total minutes spent on task, mutual information of best feature with target on the test set, and total global feature importance assigned by AutoML service on development set, for participants of varying experience sorted by type of background.}
  \label{fig:bg_exp_perf_summary}
  \Description{Figure 4 includes 12 subplots showing the distribution (represented by a box plot) of four dependent variables (TLX - Overall, Minutes Spent, Mutual Information, and Feature Importance) against three independent variables (Data Science, Software Development and Domain Expertise) each broken down into three levels (Beginner, Intermediate, and Expert). Comparing the three levels shown in each of the subplots, most are similar statistically. Data science background across levels has little relationship with the dependent outcomes. But software development background and domain expertise are positive correlated with mutual information with increasing level.}
\end{figure}

In answering this question, we look at six dimensions of participants' backgrounds. Because many are complementary, for purposes of analysis, we collapse them into the broader categories of data science background, software development background, and domain expertise. Our main dependent variables for illustrating participant experience are the overall cognitive load (TLX - Overall) and total minutes spent on the task (Minutes Spent). Our main dependent variables for illustrating participant performance are two measures of the ML performance of each feature: its mutual information with the target and its feature importance as assessed by AutoML. We summarize the relationship between background, experience, and performance measures in \Cref{fig:bg_exp_perf_summary}.

\textbf{Beginners find the task accessible.} Beginners found the task to be accessible, as across different backgrounds, beginners had a median task demand of 45.2 (lower is less demanding, p25=28.5, p75=60.4). The groups that found the task most demanding were those with little experience analyzing survey data or developing open-source projects.

\textbf{Experts find the task less demanding but perform similarly.} We found that broadly, participants with increased expertise in any of the background areas perceived the task as less demanding. However, data science and feature engineering experts spent more time working on the task than beginners did. They were not necessarily using this time to fix errors in their feature definitions, as they invoked the \Ballet client's validation functions fewer times, according to telemetry data (16 times for experts, 33.5 times for non-experts). They may have been spending more time learning about the project and data without writing code. Then, they may have used their preferred methods to help evaluate their features during development. However, our hypothesis that experts would onboard faster than non-experts when measured by minutes spent learning about \Ballet (a component of the total minutes spent) is rejected for data science background (Mann-Whitney U=85.0, $\Delta$ medians -6.0 minutes) and for software development background (U=103.0, $\Delta$ medians -1.5 minutes).

\textbf{Domain expertise is critical.} Of the different types of participant background, domain expertise had the strongest relationship with better participant outcomes. This is encouraging because it suggests that if collaborative data science projects attract experts in the project domain, these experts can be successful as long as they have data science and software development skills above a certain threshold and are supported by user-friendly tooling like \Assemble. One explanation for the relative importance of domain expertise is that participants can become overwhelmed or confused by \textit{dataset challenges} with the wide and dirty survey dataset: \textit{``There are a lot of values in the data, and I couldn't figure out the meaning of the values, because I didn't know much about the topic''} (P20). We speculate that given the time constraints of the task, participants who were more familiar with survey data analysis were able to allocate time they would have spent here to learning about \Ballet or developing features. We find that beginners spent substantially more time learning about the prediction problem and data --- a median of 36 minutes  vs. 13.6 minutes for intermediate and expert participants (Mann-Whitney U=36.5, p=0.064, $n_1$=6, $n_2$=21).

\subsection{RQ3: Collaborative feature engineering code}
\label{sec:results:collaborative-feature-engineering-code}

We were interested in understanding the kind of feature engineering code that participants write in this collaborative setting.
Participants in the case study contributed 28 feature definitions to the project, which together extract 32 feature values. The features had 47 transformers, with most \featurefunctions applying a single transformer to the input but some applying up to four transformers sequentially.

\textbf{Feature engineering primitives.} Participants collectively used 10 different feature engineering primitives (\Cref{app:fteng:feature-engineering-primitives}). Our source code analysis shows that 17/47 transformers were \inline{FunctionTransformer} primitives that can wrap standard statistical functions or are used by \Ballet to automatically wrap anonymous functions. Use of these was broadly split between simple functions to process variables that needed minimal cleaning/transformation vs. complex functions that extracted custom mappings from ordinal or categorical variables based on a careful reading of the survey codebook.

\textbf{Feature characteristics.} These \featurefunctions consumed 137 distinct variables from the raw ACS responses, out of a total of 494 present in the entities table. Most of these variables were consumed by just one feature, but several were transformed in different ways, such as \inline{SCHL} (educational attainment), which was an input to five different features. Thus 357 variables, or 72\%, were ignored by the collaborators. Some were ignored because they are not predictive of personal income. For example, the end-to-end AutoML model that operates directly on the raw ACS responses assigns a feature importance of less than $0.001$ to 418 variables (where the feature importance values sum to one). However, there may still be missed opportunities by the collaborators, as the AutoML model assigns feature importance of greater than $0.01$ to seven variables that were not used by any of the participants' features --- such as \inline{RELP}, which indicates the person's relationship to the ``reference person'' in the household and is an intuitive predictor of income because it allows the modeler to differentiate between adults who are dependents of their parents. This suggests an opportunity for developers of collaborative frameworks like \Ballet to provide more formal direction about where to invest feature engineering effort --- for example, by providing methods to summarize the inputs that have or have not been included in patches, in line with the \emph{distribution of work} theme that emerged from participant responses and the challenge of task management. Of the features, 11/28 had a learned transformer while the remainder did not learn any feature engineering-specific parameters from the training data, and 14/47 transformers were learned transformers.

\textbf{Feature definition abstraction.} The \inline{Feature} abstraction of \Ballet yields a one-to-one correspondence between the task and feature definitions. This new programming paradigm required participants to adjust their usual feature engineering toolbox. For many respondents, this was a positive change, with benefits for reusability, shareability, and tracking prior features: \textit{``It allows for a better level of abstraction as it raises Features up to their own entity instead of just being a standalone column''} (P16). For others, it was difficult to adjust, and participants noted challenges in learning how to express their ideas using transformers and feature engineering primitives and how to debug failures.

\subsection{RQ4: Comparative performance}
\label{sec:results:comparative-performance}

\begin{table}

  \caption{ML Performance of \Ballet and alternatives. The AutoML feature engineering is not robust to changes from the development set and fails with errors on almost half of the test rows. But when using the feature definitions produced by the \Ballet collaboration, the AutoML method outperforms human experts.}    \label{tab:comparative_performance}

      \begin{tabular}{ll|ccccc}
      \toprule
      \makecell{Feature \\ Engineering} & Modeling & Accuracy & Precision & Recall & F1 & \makecell{Failure \\ rate} \\
      \midrule
      \Ballet & AutoML & \textbf{0.876} & \textbf{0.838} & 0.830 & \textbf{0.834} & 0.000 \\
      AutoML & AutoML & 0.462 & 0.440 & 0.423 & 0.431 & 0.475 \\
      \Ballet & Expert 1 & 0.828 & 0.799 & 0.707 & 0.734 & 0.000 \\
      \Ballet & Expert 2 & 0.840 & 0.793 & 0.858 & 0.811 & 0.000 \\
      Expert 1 & Expert 1 & 0.814 & 0.775 & 0.686 & 0.710 & 0.000 \\
      Expert 2 & Expert 2 & 0.857 & 0.809 & \textbf{0.867} & 0.828 & 0.000 \\
      \bottomrule
    \end{tabular}  
\end{table}

While we focus on better understanding how data science developers work together in a collaborative setting, ultimately one important measure of the success of a collaborative model is its ability to demonstrate good ML performance. To evaluate this, we compare the performance of the feature engineering pipeline built by the case study participants against several alternatives built from our baseline solutions we obtained from outside data science experts and a commercial AutoML service, Google Cloud AutoML Tables (\Cref{sec:user-studies:case-study}).

We found that among these alternatives, the best ML performance came from using the \Ballet feature engineering pipeline and passing the extracted feature values to AutoML Tables (\Cref{tab:comparative_performance}). This hybrid human-AI approach outperformed end-to-end AutoML Tables and both of the outside experts. This finding also confirms previous results suggesting that feature engineering is sometimes difficult to automate, and that advances in AutoML have led to expert- or super-expert performance on clean, well-defined inputs.
\textbf{Qualitative differences.} The three approaches to the task varied widely. Due to \Ballet{'s} structure, participants spent all of their development effort on creating a small set of high-quality features. AutoML Tables performs basic feature engineering according to the inferred variable type (normalize and bucketize numeric variables, create one-hot encoding and embeddings for categorical variables) but spends most of its runtime budget searching and tuning models, resulting in a gradient-boosted decision tree for solving the census problem. The experts similarly performed minimal feature engineering (encoding and imputing); the resulting models were a minority class oversampling step followed by a tuned AdaBoost classifier (Expert 1) and a custom greedy forward feature selection step followed by a linear probability model (Expert 2).

\section{Discussion}
\label{sec:discussion}

In this section, we reflect on Ballet and on the case study presented in this paper with a particular eye toward scale, human factors, security, privacy, and the limitations of our research. We also discuss future areas of focus for HCI researchers and designers of collaborative frameworks.

\subsection{Effects of scale}
\label{sec:discussion:effects-of-scale}

Although Ballet makes strides in scaling collaborative data science projects beyond small teams, we expect additional challenges to arise as collaborations are scaled even further.

The number of possible direct communication channels in a project scales quadratically with the number of developers \cite{brooks1995mythical}. At the scale of our case study, communication among collaborators can take place effectively through discussion threads, chat rooms, and shared work products. But projects with hundreds or thousands of developers require other strategies, such as search, filtering, and recommendation of relevant content. The metadata exposed by feature definitions (\Cref{sec:fteng:feature-definitions}) makes it possible to explore these functionalities in future work.

A task management challenge that goes hand-in-hand with communication is distribution of work, a theme from our qualitative analysis in \Cref{sec:results}. Even at the scale of our case study, some collaborators wanted \Ballet itself to support them in the distribution of feature engineering work. At larger scales, this need becomes more pressing if redundant work is to be avoided. In addition to strategies like partitioning the variable space and surfacing unused variables for developers, other solutions may include ticketing systems, clustering of variables and partitioning of clusters, and algorithms to rank variables by their impact after transformations have been applied.

\subsection{Effects of culture}
\label{sec:discussion:effects-of-culture}

Data science teams are often made up of like-minded people from similar backgrounds. For example, \citet{choi2017characteristics} report that most of the open data analysis projects they reviewed were comprised of people who already knew each other --- partly because teammates wanted to be confident that everyone there had the required expertise.

Our more formalized and structured notion of collaboration may allow data science developers with few or no personal connections to form more diverse, cross-cultural teams. For example, the \pci project included collaborators from four different global regions (North America, Europe, Asia, and the Middle East). The support for validating contributions like feature definitions with statistical and software quality measures may allow teammates to have confidence in each other even without knowing each other's backgrounds or specific expertise.

\subsection{Security}
\label{sec:discussion:security}

Any software project that receives untrusted code must be mindful of security considerations. The primary threat model for \Ballet projects is a well-meaning collaborator that submits poorly-performing feature definitions or inadvertently ``breaks'' the feature engineering pipeline, a consideration that partly informed the acceptance procedure in \Cref{sec:fteng:testing}. \Ballet{'s} support for automatic merging of accepted features presents a risk that harmful code may be embedded within otherwise relevant features. While requiring a maintainer to give final approval for accepted features is a practical defense, defending against malicious contributions is an ongoing struggle for open-source projects \cite{payne2002security, decan2016topology, npm4}.

\subsection{Privacy}
\label{sec:discussion:privacy}

Data is, to no surprise, central to data science. This can pose challenges for open data projects if the data they want to use is sensitive or confidential --- for example, if it contains personally identifiable information. The main way to address this issue is to secure the dataset but open the codebase. In this formulation, access to the data is limited to those who have undergone training or signed a restricted use agreement. But at the same time, the entirety of the code, including the feature definitions, can be developed publicly without revealing any non-public information about the data. With this strategy, developers and maintainers must monitor submissions to ensure that data is not accidentally copied into source code files --- a process that can be automated, similar to scanning for secure tokens and credentials \cite{meli2019how, glanz2020hidden}.

One alternative is to make the entire repository private, ensuring that only people who have been approved have access to the code and data. However, this curtails most of the benefits of open data science and makes it more difficult to attract collaborators.

Another alternative is to anonymize data or use synthetic data for development while keeping the actual data private and secure. Recent advances in synthetic data generation \cite{xu2019modeling, patki2016synthetic} allow a synthetic dataset to be generated with the same schema and joint distribution as the real dataset, even for complex tables. This may be sufficient to allow data science developers to discover patterns and create feature definitions that can then be executed on the real, unseen dataset. This follows work releasing sensitive datasets for analysis in a privacy-preserving way using techniques like differential privacy \cite{dwork2008differential}. Indeed, the U.S. Census is now using differential privacy techniques in the release of data products such as the ACS \citep{abowd2020modernization}. Analyses developed in an open setting could then be re-run privately on the original data according to the privacy budget of each researcher.

\subsection{Interpretability and documentation}
\label{sec:discussion:interpretability-and-documentation}

\citet{zhang2020how} observe that data science workers rarely create documentation about their work during feature engineering, suggesting that human decision-making may be reflected in the data pipeline while ``simultaneously becoming invisible.'' This poses a risk for replicability, maintenance, and interpretability. In \Ballet, the structure provided by our feature definition abstractions means that the resulting feature values have clear provenance and are interpretable, in the sense that for each column in the feature matrix, the raw variables used and the exact transformations applied are easily surfaced and understood. Feature value names (columns in the feature matrix) can be provided by data scientists when they create feature definitions, or reasonable names can be inferred by Ballet from the available metadata and through lightweight program analysis.

\subsection{Feature maintenance}
\label{sec:discussion:feature-maintenance}

Just as software libraries require maintenance to fix bugs and make updates in response to changing APIs or dependencies, so too do feature definitions and feature engineering pipelines. Feature maintenance may be required in several situations. First, components from libraries used in a feature definition, such as the name or behavior of an imputation primitive, could change. Second, the schema of the target dataset could change, such as if a survey is conducted in a new year, with certain questions from prior years replaced with new ones.\footnote{For example, the U.S. Census has modified the language used to ask about respondents' race several times in response to an evolving understanding of this construct. A changelog \citep{acs2018_pums_readme} of a recently conducted survey compared to the prior year contained 42 entries.} Third, feature maintenance may be required due to distribution shift, in which new observations following the same schema have a different data distribution, causing the assumptions reflected in a feature definition to be invalidated.

Though we have focused mainly on the scale of a collaboration in terms of the number of code contributors, another important measure of scale is the length of time the project remains in a developed, maintained state, and as such is useful to consumers. As projects age, these secondary issues of feature maintenance, as well as dataset and model versioning and changing usage scenarios, become more salient.

A similar development workflow to the one presented in this paper could also be used for feature maintenance, and researchers have pointed out that the open-source model is particularly well suited for ensuring software is maintained \cite{johnson2006collaboration}. Currently, \Ballet focuses on supporting the addition of new features; to support the modification of existing features would require additional design considerations, such as how developers using \Assemble could indicate which feature should be updated/removed by their pull request. Automatically detecting the need for maintenance due to distribution shift or otherwise is an important research direction, and can be supported in the meantime by \textit{ad hoc} statistical tests created by project maintainers.

\subsection{Ethical considerations}

As the field of machine learning rapidly advances, more and more ethical considerations are being raised, including of recent models \cite{bender2021dangers, bolukbasi2016man, strubell2019energy}. The same set of concerns could also be raised about any model developed using \Ballet. Addressing the underlying issues is beyond the scope of this paper. However, we emphasize that \Ballet provides several benefits from an ethical perspective. The open-source setting means that models are open and transparent from the outset. Similarly, we focus on the development of models that aim to address societal problems, such as vehicle fatality prediction and government survey optimization.

\subsection{Limitations}
\label{sec:discussion:limitations}

There are several limitations to our approach. Feature engineering is a complex process, and we have not yet provided support for several common practices (or potential new practices). For example, many features are trivial to specify and can be enumerated by automated approaches \cite{kanter2015deep}, and some data cleaning and preparation can be performed automatically. We have referred the responsibility for adding these techniques to the feature engineering pipeline to individual project maintainers. Similarly, feature engineering with higher-level features, that operate on variable types rather than specific variables or that operate on existing feature values rather than raw variables, could enhance developer productivity.

Feature engineering is only one part of the larger data science process, albeit an important one. Indeed, many domains, including computer vision and natural language processing, have largely replaced manually engineered features with learned ones extracted by deep neural networks. Applying our conceptual framework to other aspects of data science, like data programming or ensembling in developing predictive models, can increase the impact of collaborations. Similarly, improving collaboration in other aspects of data work --- like data journalism, exploratory data analysis, causal modeling, and neural network architecture design --- remains an important challenge.

\section{Conclusion}
\label{sec:conclusion}

We provided a conceptual framework for collaboration in open-source data science projects and implemented it in \Ballet. Our work develops the conceptual, algorithmic, engineering, and interaction approaches to move forward the vision of large-scale, collaborative data science. The success of the \pci project shows the potential of this approach and gives further direction for framework developers.

\makeatletter
\if@ACM@anonymous
	\else
	
\begin{acks}
We'd like to thank the following people: members of the Data To AI Lab for feedback and pilot testing, participants in the \textit{predict-house-prices} demonstration at MLSys 2020, participants in the \textit{predict-census-income} case study, Ignacio Arnaldo, Zhuofan Xie, and Fahad Alhasoun. This work is supported in part by NSF Award 1761812.
\end{acks}

\fi
\makeatother

\clearpage

\bibliographystyle{ACM-Reference-Format}
\bibliography{references}

\par\bigskip\noindent\small\normalfont{Received January 2021; revised April 2021; revised July 2021; accepted July 2021}\par

\clearpage

\nobalance
\appendix

\section{Largest collaborations}
\label{app:largest-collaborations}

In \Cref{tab:project-sizes}, we show the number of unique contributors to selected large open-source collaborations in either software engineering or ML modeling, as of October 2020. In this section, we briefly describe the systematic review of GitHub projects that underlies this table.

We obtain a list of large software engineering collaborations as follows. We define a software engineering collaboration as a source code repository for a software library, framework, or application. We use GitHub's API to query for repositories with at least 50 forks and 250 stars. We exclude repositories that have the topics \inline{awesome} or \inline{interview} as these represent large crowd-sourced wikis/resources that are not actually software projects. For each repository, we count the number of contributors by scraping the count from the ``Contributors'' section of the repository home page. As computed by GitHub, this represents the number of unique users (including developers whose emails appear in the git log but do not have an associated GitHub account) who have at least one commit, excluding merge commits and empty commits. In the case that the scrape fails, we fall back to using the GitHub API ``List repository contributors'' endpoint and count the number of results.

We obtain a list of large ML modeling collaborations as follows. First, we define an ML modeling collaboration as a source code repository for a predictive ML model. We do not impose any requirements on dataset availability. We first search for projects with topics including \inline{kaggle-competition} and \inline{tensorflow-model}, for projects that mention the term \inline{cookiecutterdatascience} in the README, and filter to results with greater than five forks and zero stars. We also searched for projects created by well-known organizations like Google Research, Intel, IBM, and NVIDIA. Then, we augment this list with an additional list of data science projects of interest that we manually curate as part of our research (32 in total). This list includes projects from organizations such as dssg (Data Science for Social Good), Data For Democracy, and Microsoft's AI for Earth initiative. We use the same methodology to count contributions as in the case of software projects.

There are some limitations to our methodology for identifying large data science collaborations. There may be additional data science search terms or projects that we did not find. And we purposefully excluded projects that defined many different models in subdirectories, so-called ``model zoos'' or ``model gardens.'' In these projects, each subdirectory is a different model contributed by a different set of contributors, and counting contributions per model is a non-trivial task. For example, \inline{tensorflow/models} has 709 contributors but at least 50 model implementations. Similarly, we excluded projects that are frameworks for applying one of many possible models, such as the \inline{OpenNMT} framework for machine translation or Facebook Research's \inline{detectron2} platform for object detection. Furthermore, this represents a narrow view of contribution that excludes creating and triaging bug reports, reviewing code, and creating documentation outside of the source tree.

Replication files are available here: \url{\replicationurl}.

\section{A language for feature engineering}
\label{app:fteng}

In this section, we provide more detail on feature engineering within \Ballet. Additional details are also presented in \citet[Chapter~3]{smith2021collaborative}.

We start from the insight that feature engineering can be represented as a dataflow graph over individual features. We structure code that extracts a group of feature values as a patch, calling these \textit{feature definitions} and representing them with a \inline{Feature} interface. Feature definitions are composed into a feature engineering pipeline product. Newly contributed feature definitions are accepted if they pass a two-stage acceptance procedure that tests both the feature API and its contribution to ML performance. Finally, the plugin specifies the organization of modules within a repository to allow features to be collected programmatically.

In Ballet, we create a flexible and powerful language for feature engineering that is embedded within the larger framework. It supports functionality such as learned feature transformations, supervised feature transformations, nested transformer steps, syntactic sugar for functional transformations, data frame-style transformations, and recovery from errors due to type incompatibility.

\subsection{Feature definitions}
\label{app:fteng:feature-definitions}

A \textit{feature definition} is the code that is used to extract semantically related feature values from raw data. Let us observe data instances ${\D = (\mathbf{v}_i, \mathbf{y}_i)_{i=1}^{N}}$, where $\mathbf{v}_i \in \V$ are the raw variables and $\mathbf{y}_i \in \Y$ is the target. In this formulation, the raw variable domain $\V$ includes strings, missing values, categories, and other non-numeric types that cannot typically be inputted to learning algorithms. Thus our goal in feature engineering is to develop a learned map from $\V$ to $\X$ where $\X \subseteq \R^n$ is a real-valued feature space.

\begin{definition}
A \emph{\featurefunction} is a learned map from raw variables in one data instance to feature values, $f: (\V, \Y) \to \V \to \X$.
\end{definition}

We indicate the map learned from a specific dataset $\D$ by $f^\D$, \ie $f^D(v) = f(D)(v)$.

A \featurefunction can produce output of different dimensionality. Let $q(f)$ be the dimensionality of the feature space $\X$ for a feature $f$. We call $f$ a \emph{scalar-valued feature} if $q(f) = 1$ or a \emph{vector-valued feature} if $q(f) > 1$. For example, the embedding of a categorical variable, such as a one-hot encoding, would result in a vector-valued feature.

We can decompose a \featurefunction into two parts, its \emph{input projection} and its \emph{transformer steps}. The input projection is the subspace of the variable space that it operates on, and the transformer steps, when composed together, equal the learned map on this subspace.

\begin{definition}
  A \emph{feature input projection} is a projection from the full variable space to the \emph{feature input space}, the set of variables that are used in a \featurefunction, $\pi: \V \to \V$.
\end{definition}

\begin{definition}
  A \emph{feature transformer step} is a learned map from the variable space to the variable space, $f_i: (\V, \Y) \to \V \to \V$.
\end{definition}

We require that individual feature transformer steps compose together to yield the \featurefunction, where the first step applies the input projection and the last step maps to $\X$ rather than $\V$. That is, each transformer step applies some arbitrary transformation as long as the final step maps to allowed feature values.

\begin{align*}
  f_0&: (V, Y) \mapsto \pi \\
  f_i \circ f_{i-1}&: (V, Y) \mapsto f_i(f_{i-1}(\dots, Y)(V), Y) \\
  f_n&: (\V, \Y) \to \V \to \X \\
  f&: (V, Y) \mapsto f_n(\dots(f_1(f_0(V, Y)(V), Y)\dots), Y)
\end{align*}

The \inline{Feature} class in Ballet is a way to express a \featurefunction in code. It is a tuple (\inline{input}, \inline{transformer}). The input declares the variable(s) from $\V$ that are needed by the feature, which will be passed to \inline{transformer}, one or more transformer steps. Each transformer step implements the learned map via \inline{fit} and \inline{transform} methods, a standard interface in machine learning pipelines \cite{buitinck2013api}. A data science developer then simply provides values for the input and transformer of a \inline{Feature} object in their code. Additional metadata, including \inline{name}, \inline{description}, \inline{output}, and \inline{source}, is also exposed by the \inline{Feature} abstraction. For example, the feature \inline{output} is a column name (or set of column names) for the feature value (or feature values) in the resulting feature matrix; if it is not provided by the developer, it can be inferred by Ballet using various heuristics.

Two example feature definitions are shown in \Cref{fig:feature-examples}.

\subsection{Learned feature transformations}
\label{app:fteng:learned-feature-transformations}

In machine learning, we estimate the generalization performance of a model by evaluating it on a set of test observations that are unseen by the model during training. \emph{Leakage} is a problem in which information about the test set is accidentally exposed to the model during training, artificially inflating its performance on the test set and thus underestimating generalization error.

Each \featurefunction learns a specific map $\V \to \X$ from $\D$, such that any parameters it uses, such as variable means and variances, are learned from the development (training) dataset. This formalizes the separation between development and testing data to avoid any leakage of information during the feature engineering process.

As in the common pattern, a feature engineering pipeline by itself or within a model has two stages within training: a fit stage and a transform stage. During the fit stage, parameters are learned from the training data and stored within the individual transformer steps. During the transform stage, the learned parameters are used to transform the raw variables into a feature matrix. The same parameters are also used at prediction time.

\subsection{Nested feature definitions}
\label{app:fteng:nested-feature-definitions}

Feature definitions or \featurefunctions can also be nested within the \inline{transformer} field of another feature. If an existing feature is used as one transformer step in another feature, when the new \featurefunction is executed, the nested feature is executed in a sub-procedure and the resulting feature values are available to the new feature for further transformation. Data science developers can also introspect an existing feature to access its own \inline{input} and \inline{transformer} attributes and use them directly within a new feature.

Through its support for nested feature definitions, Ballet allows collaborating developers to define an arbitrary directed acyclic dataflow graph from the raw variables to the feature matrix.

\subsection{Feature engineering primitives}
\label{app:fteng:feature-engineering-primitives}

Many features exhibit common patterns, such as scaling or imputing variables using simple procedures. And while some features are relatively simple and have no learned parameters, others are more complicated to express in a fit/transform style. Data science developers commonly extract these more advanced features by manipulating development and test tables directly using popular data frame libraries, often leading to leakage. In preliminary studies (\Cref{sec:user-studies:preliminary-studies}), we found that data science developers sometimes struggled to create features one at a time, given their familiarity with writing long processing scripts. Responding to this feedback, we provided a library of \textit{feature engineering primitives} that implements many common utilities and learned transformations.

\begin{definition}
A \emph{feature engineering primitive} is a class that can be instantiated within a sequence of transformer steps to express a common feature engineering pattern.
\end{definition}

This library, \inline{ballet.eng}, includes primitives like \inline{ConditionalTransformer}, which applies a secondary transformation depending on whether a condition is satisfied on the development data, and \\ \inline{GroupwiseTransformer}, which learns a transformer separately for each group of a group-by aggregation on the development set. We also organize and re-export 77 primitives\footnote{As of ballet v0.19.} from six popular Python libraries for feature engineering, such as scikit-learn's \inline{SimpleImputer} (\Cref{tab:fteng-primitives}).

\begin{table}[th]
    \centering
    \caption[Feature engineering primitives implemented or re-exported.]{Feature engineering primitives implemented or re-exported in \inline{ballet.eng}, by library.}
    \label{tab:fteng-primitives}
    \begin{tabular}{lc}
        \toprule
        Library & Number of primitives \\
        \midrule
        \inline{ballet} & 16 \\
        \inline{category\_encoders} & 17 \\
        \inline{feature\_engine} & 29 \\
        \inline{featuretools} & 1 \\
        \inline{skits} & 10 \\
        \inline{scikit-learn} & 19 \\
        \inline{tsfresh} & 1 \\
        \bottomrule
    \end{tabular}
\end{table}

\subsection{Feature engineering pipelines}
\label{app:fteng:feature-engineering-pipelines}

Features are then composed together in a feature engineering pipeline.

\begin{definition}
Let $f_1, \dots, f_m$ be a collection of \featurefunctions, $\D$ be a development dataset, and $\D' = (V', Y')$ be a collection of new data instances.  A \emph{feature engineering pipeline} $\F = \{f_i\}_{i=1}^m$ applies each \featurefunction to the new data instances and concatenates the result, yielding the feature matrix

\[
X = \F^\D(V') = f_1^\D(V') \oplus \dots \oplus f_m^\D(V').
\]

\end{definition}

A feature engineering pipeline can be thought of as similar to a collection of \featurefunctions. It is implemented in Ballet in the \inline{FeatureEngineeringPipeline} class.

In \Ballet's standard configuration, only \featurefunctions that have been explicitly defined by data science developers (and accepted by the feature validation) are included in the feature engineering pipeline. Thus, raw variables are not outputted by the pipeline unless they are explicitly requested (\ie by a developer developing a feature definition that applies the identity transformation to a raw variable). In alternative configurations, project maintainers can define a set of fixed feature definitions to include in the pipeline, which can include a set of important raw variables with the identity transformation or other transformations applied.

\subsection{Feature execution engine}
\label{app:fteng:feature-execution-engine}

Ballet's feature execution engine is responsible for applying the full feature engineering pipeline or an individual feature to extract feature values from a given set of data instances. Each \featurefunction within the pipeline is passed the input columns it requires, which it then transforms appropriately, internally using one or more transformer steps (\Cref{fig:feature-engineering-pipeline-example}). It operates as follows, starting from a set of \inline{Feature} objects.

\begin{enumerate}

\item The transformer steps of each feature are postprocessed in a single step. First, any syntactic sugar is replaced with the appropriate objects. For example, an anonymous function is replaced by a \inline{FunctionTransformer} object that applies the function, or a tuple of an input and another transformer is replaced by an \inline{SubsetTransformer} object that applies the transformer on the given subset of the input and passes through the remaining columns unchanged. Second, the transformer steps are all wrapped in functionality that allows them to recover from any errors that are due to type incompatibility. For example, if the underlying transformation expects a 1-d array (column vector), but receives as input a 2-d array with a single column, the wrapper will catch the error, convert the 2-d array to a 1-d array, and retry the transformation. The wrapper pre-defines a set of these ``conversion approaches'' (one of which is the identity transformation), which will be tried in sequence until one is successful. The successful approach is stored so that it can be re-used during subsequent applications of the feature.

\item The features are composed together into a feature engineering pipeline object.

\item The fit stage of the feature engineering pipeline is executed. For each feature, the execution engine indexes out the declared input columns from the raw data and passes them to the wrapped \inline{fit} method of the feature's transformer. (This stage only occurs during training.)

\item The transform stage of the feature engineering pipeline is executed. For each feature, the execution engine indexes out the declared input columns from the raw data and passes them to the wrapped \inline{transform} method of the feature's transformer.

\end{enumerate}

This process can also be parallelized across features. Since support for nested feature definitions (\Cref{app:fteng:nested-feature-definitions}) means that if features were executed independently there may be redundant computation if there were dependencies between features, this would necessitate a more careful approach in which the features are first sorted topologically and then resulting feature values are cached after first computation. For very large feature sets or datasets, full-featured dataflow engines should be considered.

\subsection{Acceptance procedures for feature definitions}
\label{app:fteng:testing}

Contributions of feature engineering code, just like other code contributions, must be evaluated for quality before being accepted in order to mitigate the risk of introducing errors, malicious behavior, or design flaws. For example, a \featurefunction that produces non-numeric values can result in an unusable feature engineering pipeline. Large feature engineering collaborations can also be susceptible to ``feature spam,'' a high volume of low-quality feature definitions (submitted either intentionally or unintentionally) that harm the collaboration \cite{smith2017featurehub}. Modeling performance can suffer and require an additional feature selection step --- violating the working pipeline invariant --- and the experience of other collaborators can be harmed if they are not able to assume that existing feature definitions are high-quality.

To address these possibilities, we extensively validate feature definition contributions for software quality and ML performance. Validation is implemented as a test suite that is both exposed by the Ballet client and executed in CI for every pull request. Thus, the same method that is used in CI for validating feature contributions is available to data science developers for debugging and performance evaluation in their development environment. Ballet Bot can automatically merge pull requests corresponding to accepted feature definitions and close pull requests corresponding to rejected feature definitions.

This automatic acceptance procedure is defined for the addition of new feature definitions only, while acceptance procedures for edits or deletions of feature definitions is important future work.

\subsubsection{Feature API validation}
\label{app:fteng:testing:feature-testing}

User-contributed feature definitions should satisfy the \inline{Feature} interface and successfully deal with common error situations, such as intermediate computations producing missing values. We fit the \featurefunction to a separate subsampled training dataset in an isolated environment and extract feature values from subsampled training and validation datasets, failing immediately on any implementation errors. We then conduct a battery of 15 tests to increase confidence that the \featurefunction would also extract acceptable feature values on unseen inputs (\Cref{tab:test-suite-feature-api-validation}). Each test is paired with ``advice'' that can be surfaced back to the user to fix any issues (\Cref{fig:overview}).

\begin{table}
  \resizebox{\linewidth}{!}{    {\small\fontfamily{cmtt}\selectfont
    \begin{tabular}{lll}
      \toprule
      IsFeatureCheck &
      HasCorrectInputTypeCheck &
      HasCorrectOutputDimensionsCheck \\
      HasTransformerInterfaceCheck &
      CanFitCheck &
      CanFitOneRowCheck \\
      CanTransformCheck &
      CanTransformNewRowsCheck &
      CanTransformOneRowCheck \\
      CanFitTransformCheck &
      CanMakeMapperCheck &
      NoMissingValuesCheck \\
      NoInfiniteValuesCheck &
      CanDeepcopyCheck &
      CanPickleCheck \\
      \bottomrule
    \end{tabular}    }
  }
  \caption{Feature API validation suite in (\inline{ballet.validation.feature\_api.checks}) that ensures the proper functioning of the shared feature engineering pipeline.}
  \label{tab:test-suite-feature-api-validation}
\end{table}

Another part of feature API validation is an analysis of the changes introduced in a proposed pull request to ensure that the required project structure is preserved and that the collaborator has not accidentally included irrelevant code that would need to be evaluated separately.\footnote{This ``project structure validation'' is only relevant in CI and is not exposed by the Ballet client.} A feature contribution is valid if it consists of the addition of a valid source file within the project's \inline{src/features/contrib} subdirectory that also follows a specified naming convention using the user's login name and the given feature name. The introduced module must define exactly one object --- an instance of \inline{Feature} --- which will then be imported by the framework.

\subsubsection{ML performance validation}
\label{app:fteng:testing:ml-performance-validation}

A complementary aspect of the acceptance procedure is validating a feature contribution in terms of its impact on machine learning performance, which we cast as a streaming feature definition selection (SFDS) problem. This is a variant of streaming feature selection where we select from among feature definitions rather than feature values. Features that improve ML performance will pass this step; otherwise, the contribution will be rejected. Not only does this discourage low-quality contributions, but it provides a way for collaborators to evaluate their performance, incentivizing more deliberate and creative feature engineering.

We first compile requirements for an SFDS algorithm to be deployed in our setting, including that the algorithm should be stateless, support real-world data types (mixed discrete and continuous), and be robust to over-submission. While there has been a wealth of research into streaming feature selection \cite{zhou2005streaming, wu2013online, wang2015online, yu2016scalable}, no existing algorithm satisfies all requirements. Instead, we extend prior work to apply to our situation \cite{li2013group,kraskov2004estimating}. Our SFDS algorithm proceeds in two stages.\footnote{Indeed, we abbreviate the general problem of streaming feature definition selection as SFDS, and also call our algorithm to solve this problem SFDS. We trust that readers can disambiguate based on context.} In the \textit{acceptance} stage, we compute the conditional mutual information of the new feature values with the target, conditional on the existing feature matrix, and accept the feature if it is above a dynamic threshold. In the \textit{pruning} stage, existing features that have been made newly redundant by accepted features can be pruned. Details are presented in the following sections.

\subsection{Streaming feature definition selection}
\label{app:fteng:testing:streaming-feature-definition-selection}

Feature selection is a classic problem in machine learning and statistics \citep{guyon2003introduction}. The problem of feature selection is to select a subset of the available feature values such that a learning algorithm that is run on the subset generates a predictive model with the best performance according to some measure.

\begin{definition}
  The \emph{feature selection problem} is to select a subset of feature values that maximizes some utility,
  \begin{equation}
    X^* = \argmax_{X' \in \mathcal{P}(X)} U(X'),
  \end{equation}
\end{definition}

where $\mathcal{P}(A)$ denotes the power set of $A$. For example, $U$ could measure the empirical risk of a model trained on $X'$.

If there exists a group structure in $X$, then this formulation ignores the group structure and allows feature values to be subselected from within groups. In some cases, like ours, this may not be desirable, such as if it is necessary to preserve the coherence and interpretability of each feature group. In the case of feature engineering using \featurefunctions, it further conflicts with the understanding of each \featurefunction as extracting a semantically related set of feature values.

Thus we instead consider the related problem of feature definition selection.

\begin{definition}
  The \emph{feature definition selection} problem is to select a subset of feature definitions that maximizes some utility,
  \begin{equation}
    \F^* = \argmax_{\F' \in \mathcal{P}(\F)} U(\F'),
  \end{equation}
\end{definition}

This constrains the feature selection problem to select either all of or none of the feature values extracted by a given feature.

In Ballet, as collaborators develop new features, each feature arrives at the project in a streaming fashion, at which point it must be accepted or rejected immediately. Streaming feature definition selection is a streaming extension of feature definition selection.

\begin{definition}
  Let $\Gamma$ be a feature definition stream of unknown size, let $\F$ be the set of features accepted as of some time, and let $f \in \Gamma$ arrive next. The \emph{streaming feature definition selection} problem is to select a subset of feature definitions that maximizes some utility,
  \begin{equation}
    \F^* = \argmax_{\F' \in \mathcal{P}(\F \cup f)} U(\F').
  \end{equation}
\end{definition}

Streaming feature definition selection consists of two decision problems, considered as sub-procedures. The \emph{streaming feature definition acceptance} decision problem is to \emph{accept} $f$, setting $\F \leftarrow \F \cup f$, or \emph{reject}, leaving $\F$ unchanged. The \emph{streaming feature pruning} decision problem is to remove a subset $\F_0 \subset \F$ of low-quality features, setting $\F = \F \setminus \F_0$.

\paragraph{Design criteria}
\label{app:fteng:testing:streaming-feature-definition-selection:design-criteria}

Streaming feature definition selection algorithms must be carefully designed to best support collaborations in Ballet. We consider the following design criteria, motivated by engineering challenges, security risks, and experience from system prototypes:

\begin{enumerate}

  \item \textit{Definitions, not values.} The algorithm should have first-class support for feature definitions (or feature groups) rather than selecting individual feature values.

  \item \textit{Stateless.} The algorithm should require as inputs only the current state of the Ballet project (\ie the problem data and accepted features) and the pull request details (\ie the proposed feature). Otherwise, each Ballet project (\ie its GitHub repository) would require additional infrastructure to securely store the algorithm state.

  \item \textit{Robust to over-submission.} The algorithm should be robust to processing many more feature submissions than raw variables present in the data (\ie $|\Gamma| \gg |\V|$). Otherwise malicious (or careless) contributors can automatically submit many features, unacceptably increasing the dimensionality of the resulting feature matrix.

  \item \textit{Support real-world data.} The algorithm should support mixed continuous- and discrete-valued features, common in real-world data.

\end{enumerate}

\begin{figure}[!p]
\begin{algorithm}[H]
\caption{SFDS\label{alg:sfds}}
\Input{feature stream $\Gamma$, evaluation dataset $\D$}
\Output{accepted feature set $\F$}
\BlankLine
$\F \leftarrow \varnothing$ \;
\While{$\Gamma$ {\normalfont has new features}}{
  $f \leftarrow $ get next feature from $\Gamma$ \;
  \If{$\inline{accept}(\F, f, \D)$}{
    $\F \leftarrow \inline{prune}(\F, f, \D)$ \;
    $\F \leftarrow \F \cup f$ \;
  }
}
\Return{$\F$}
\end{algorithm}

\begin{procedure}[H]
\caption{accept($\F$, $f$, $\D$)}
\Input{accepted feature set $\F$, proposed feature $f$, evaluation dataset $\D$}
\Params{penalty on number of feature definitions $\lambda_1$, penalty on number of feature values $\lambda_2$}
\Output{accept/reject}
\BlankLine
\If{$I(f(\D); Y | \F(\D)) > \lambda_1 + \lambda_2 \times q(f)$}{
  \Return{true}
}
\For{$f' \in \F$}{
  $\F' \leftarrow \F \setminus f'$ \;
  \If{$I(f(\D); Y | \F'(\D)) - I(f'(\D); Y | \F'(\D)) > \lambda_1 + \lambda_2 \times (q(f) - q(f'))$}{
    \Return{true}
  }
}
\Return{false}
\end{procedure}

\begin{procedure}[H]
\caption{prune($\F$, $f$, $\D$)}
\Input{previously accepted feature set $\F$, newly accepted feature $f$, evaluation dataset $\D$}
\Params{penalty on number of feature definitions $\lambda_1$, penalty on number of feature values $\lambda_2$}
\Output{pruned feature set $\F$}
\BlankLine
\For{$f' \in \F$}{
  $\F' \leftarrow \F \setminus f' \cup f$ \;
  \If{$I(f'(\D); Y | \F'(\D)) < \lambda_1 + \lambda_2 \times q(f')$}{
    $\F \leftarrow \F \setminus f'$ \;
  }
}
\Return{$\F$}
\end{procedure}

\caption[SFDS algorithm.]{SFDS algorithm for streaming feature definition selection. It relies on two lower-level procedures, \inline{accept} and \inline{prune} to accept new feature definitions and to possibly prune newly redundant feature definitions.}
\label{fig:sfds}
\end{figure}

\clearpage

Surprisingly, there is no existing algorithm that satisfies these design criteria. Algorithms for feature value selection might only support discrete data, algorithms for feature group selection might require persistent storage of decision parameters, etc. And the robustness criterion remains important given the results of \citet{smith2017featurehub}, in which users of a collaborative feature engineering system programmatically submitted thousands of irrelevant features, constraining modeling performance. These factors motivate us to create our own algorithm.

\paragraph{SFDS}
\label{app:fteng:testing:streaming-feature-definition-selection:sfds}

Instead, we present a new algorithm, SFDS, for streaming feature definition selection based on mutual information criteria. It extends the GFSSF algorithm \cite{li2013group} both to support feature definitions rather than feature values and to support real-world tabular datasets with a mix of continuous and discrete variables.

The algorithm (\Cref{alg:sfds}) works as follows. In the acceptance stage, we first determine if a new feature $f$ is \textit{strongly relevant}; that is, whether the information $f(\D)$ provides about $Y$ above and beyond the information that is already provided by $\F(\D)$ is above some threshold governed by hyperparameters $\lambda_1$ and $\lambda_2$, which penalize the number of features and the number of feature values, respectively. If so, we accept it immediately. Otherwise, the feature may still be \textit{weakly relevant}, in which case we consider whether $f$ and some other feature $f' \in \F$ provide similar information about $Y$. If $f$ is determined to be superior to such an $f'$, then $f$ can be accepted. Later, in the pruning stage, $f'$ and any other redundant features are pruned.

\paragraph{Alternative validators}

Maintainers of \Ballet projects are free to configure alternative ML performance validation algorithms given the needs of their own projects. While we use SFDS for the \pci project, \Ballet provides implementations of the following alternative validators: \inline{AlwaysAccepter} (accept every feature definition), \inline{MutualInformationAccepter} (accept feature definitions where the mutual information of the extracted feature values with the prediction target is above a threshold), \inline{VarianceThresholdAccepter} (accept feature definitions where the variance of each feature value is above a threshold), and \inline{CompoundAccepter} (accept feature definitions based on the conjunction or disjunction of the results of multiple underlying validators). Additional validators can be easily created by defining a subclass of \\ \inline{ballet.validation.base.FeatureAccepter} and/or \inline{ballet.validation.base.FeaturePruner}.

\end{document}